\documentclass{article}

\usepackage[preprint]{neurips_2026}

\usepackage[utf8]{inputenc} 
\usepackage[T1]{fontenc}    
\usepackage{hyperref}       
\usepackage{url}            
\usepackage{booktabs}       
\usepackage{amsfonts}       
\usepackage{nicefrac}       
\usepackage{microtype}      
\usepackage{xcolor}         

\usepackage[utf8]{inputenc} 
\usepackage[T1]{fontenc}    
\usepackage{hyperref}       
\usepackage{url}            
\usepackage{booktabs}       
\usepackage{amsfonts}       
\usepackage{nicefrac}       
\usepackage{microtype}      
\usepackage{graphicx}
\usepackage{natbib}
\usepackage{doi}
\usepackage{float}

\usepackage{minitoc}             
\usepackage{paralist}
\usepackage{enumitem}
\let\OrigAddContentsLine\addcontentsline
\renewcommand{\addcontentsline}[3]{}

\usepackage{subcaption}

\usepackage{lipsum}
\usepackage[dvipsnames]{xcolor}
\usepackage[most]{tcolorbox}

\usepackage{amsmath}
\usepackage{amssymb}
\usepackage{mathtools}
\usepackage{amsthm}
\usepackage{thmtools}
\usepackage{algorithm}
\usepackage{algorithmic}
\usepackage{listings}
\usepackage{wrapfig}
\usepackage{multirow}

\usepackage{xcolor}

\usepackage[capitalize,noabbrev]{cleveref}

\usepackage{amsmath,amsfonts,bm}

\def\eqref#1{equation~\ref{#1}}

\def\1{\bm{1}}

\DeclareMathAlphabet{\mathsfit}{\encodingdefault}{\sfdefault}{m}{sl}
\SetMathAlphabet{\mathsfit}{bold}{\encodingdefault}{\sfdefault}{bx}{n}

\usepackage{soul}
\usepackage[dvipsnames]{xcolor}

\usepackage{tikz}
\definecolor{softyellow}{RGB}{255, 249, 196}
\definecolor{softgreen}{RGB}{220, 237, 200}
\definecolor{softblue}{RGB}{227, 242, 253}
\definecolor{softpink}{RGB}{252, 228, 236}
\definecolor{forestgreen}{HTML}{228B22}

\tcbuselibrary{skins,breakable}

\newtcolorbox{qualexamplebox}[1]{
  enhanced,
  colback=black!3,
  colframe=black,
  boxrule=0.6pt,
  arc=2pt,
  left=6pt,right=6pt,top=6pt,bottom=6pt,
  title={#1}
}
\newcommand{\annblock}[3]{%
\begin{tcolorbox}[
  enhanced,
  blanker,
  sidebyside,
  sidebyside align=top,
  righthand width=28mm,
  boxrule=0pt,
  left=0pt,right=0pt,top=1pt,bottom=1pt,
  overlay={%
    \draw[#1, line width=0.6pt]
      (segmentation.north) -- (segmentation.south);
    \draw[#1, line width=0.6pt]
      (segmentation.north) -- ++(-4mm,0);
    \draw[#1, line width=0.6pt]
      (segmentation.south) -- ++(-4mm,0);

    \node[
      anchor=west,
      text=#1,
      font=\bfseries\scshape\tiny,
      align=left,
      inner sep=0pt,
      outer sep=0pt
    ] at ([xshift=1.5mm]segmentation.center) {%
      \parbox[t]{\dimexpr 28mm-6pt\relax}{#2}%
    };
  },
]
#3
\tcblower
\end{tcolorbox}\par\smallskip
}

\theoremstyle{plain}

\theoremstyle{definition}

\theoremstyle{remark}

\usepackage[textsize=tiny]{todonotes}

\usepackage{amsmath}
\usepackage{amssymb}
\usepackage{mathtools}
\usepackage{amsthm}

\usepackage[capitalize,noabbrev]{cleveref}

\title{LLM-WikiRace Benchmark: How Far Can LLMs Plan over Real-World Knowledge Graphs?}

\author{
    Juliusz Ziomek${}^{*, 1}$ , 
    William Bankes${}^{*, 2}$,
    Lorenz Wolf${}^2$,\\
    \textbf{Shyam Sundhar Ramesh${}^2$,
    Xiaohang Tang${}^2$,
    Ilija Bogunovic${}^{3,2}$}\\
    ${}^*$ Equal Contribution\\
    Correspondence to: juliusz [at] robots (dot) ox <dot> ac.uk
}

\begin{document}
\footnotetext[1]{University of Oxford, UK}
\footnotetext[2]{University College London (Centre for AI), UK}
\footnotetext[3]{University of Basel, Switzerland }

\maketitle

\begin{abstract}
  We introduce \textbf{LLM-Wikirace}, a benchmark for evaluating planning, reasoning, and world knowledge in large language models (LLMs). In LLM-Wikirace, models must efficiently navigate Wikipedia hyperlinks step by step to reach a target page from a given source, requiring look-ahead planning and the ability to reason about how concepts are connected in the real world. We evaluate a broad set of open- and closed-source models, including Gemini-3.1, GPT-5, and Claude Opus 4.6, which achieve the strongest results on the easy split of LLM-WikiRace. Performance drops sharply on hard difficulty: the best-performing model, Gemini-3.1, succeeds in only 29\% of hard games, highlighting substantial remaining challenges for frontier models. Our analysis shows that world knowledge is a necessary ingredient for success, but only up to a point, beyond this threshold, planning and long-horizon reasoning capabilities become the dominant factors. Trajectory-level analysis further reveals that even the strongest models struggle to replan after failure, frequently entering loops rather than recovering. LLM-Wikirace is a simple benchmark that reveals clear limitations in current reasoning systems, offering an open arena where planning-capable LLMs still have much to prove. Our code, data and leader board are publicly available at \url{https://llmwikirace.github.io}
\end{abstract}

\section{Introduction}
\label{sec:Introduction}
\doparttoc 
\faketableofcontents 

Large Language Models have shown strong performance on a range of complex planning and reasoning tasks \cite{cobbe2021gsm8k,hendrycksmath2021}. While such successes are typically attributed to post-training approaches \cite{yu2025dapo,guo2025deepseek}, these could not succeed without the model’s ability to leverage a large breadth of knowledge imparted through pre-training on diverse textual corpora \cite{gao2020pile}, including encyclopedic sources such as Wikipedia.

Many existing benchmarks fail to test how well models can leverage this breadth of knowledge focusing instead on narrow domains such as mathematical reasoning \cite{cobbe2021gsm8k,hendrycksmath2021}, or specific real-world tasks such as trip planning \cite{xie2024travelplanner, zheng2024natural}, and website navigation \cite{yoran2024assistantbench,zhou2023webarena}. Benchmarks that do test for breadth of knowledge \cite{yang2018hotpotqa, hendrycks2021ethics, hendryckstest2021} do not require the model to \emph{actively deploy} world knowledge to make decisions or plan. Reasoning and world knowledge have been studied in isolation, yet many real-world tasks require their tight integration. For tasks such as open-domain problem solving and autonomous research assistants the ability to recall and connect diverse pieces of knowledge is essential. In such settings, failures often arise not from a lack of reasoning ability alone but from an inability to retrieve or appropriately apply relevant world knowledge during planning. This leaves a key question: \emph{to what extent can LLMs leverage pretrained world knowledge to reason and plan in environments which require a large breadth of knowledge?}

\begin{figure*}[t!]
    \centering
    \begin{subfigure}{0.6\linewidth}
        \centering
        \includegraphics[width=\linewidth, trim=5cm 2cm 3cm 4cm, clip]{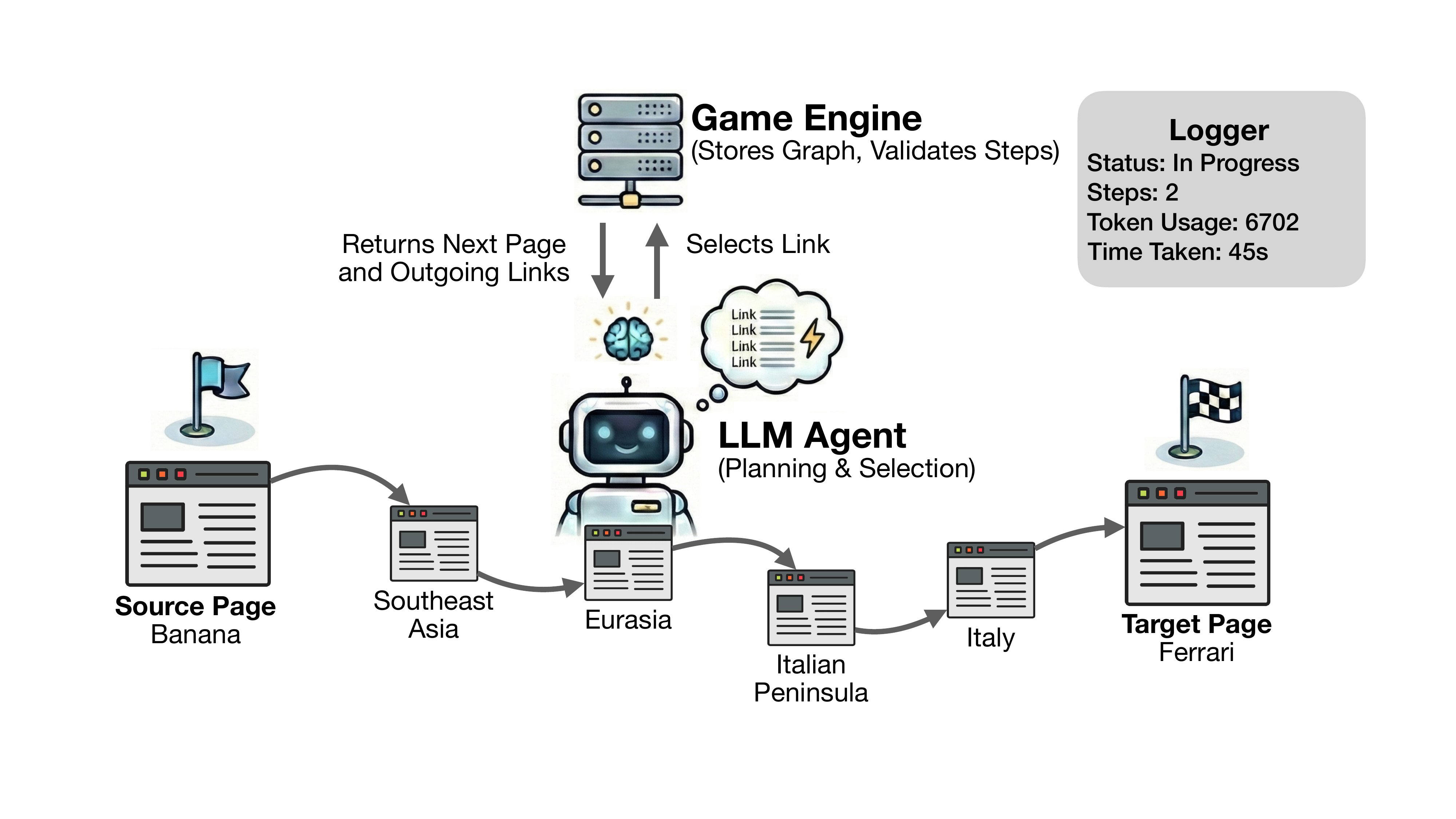}
        \caption{LLM-WikiRace setup}
        \label{fig:WikiRace_illustration}
    \end{subfigure}\hfill
    \begin{subfigure}{0.38\linewidth}
        \centering
        \includegraphics[width=\linewidth]{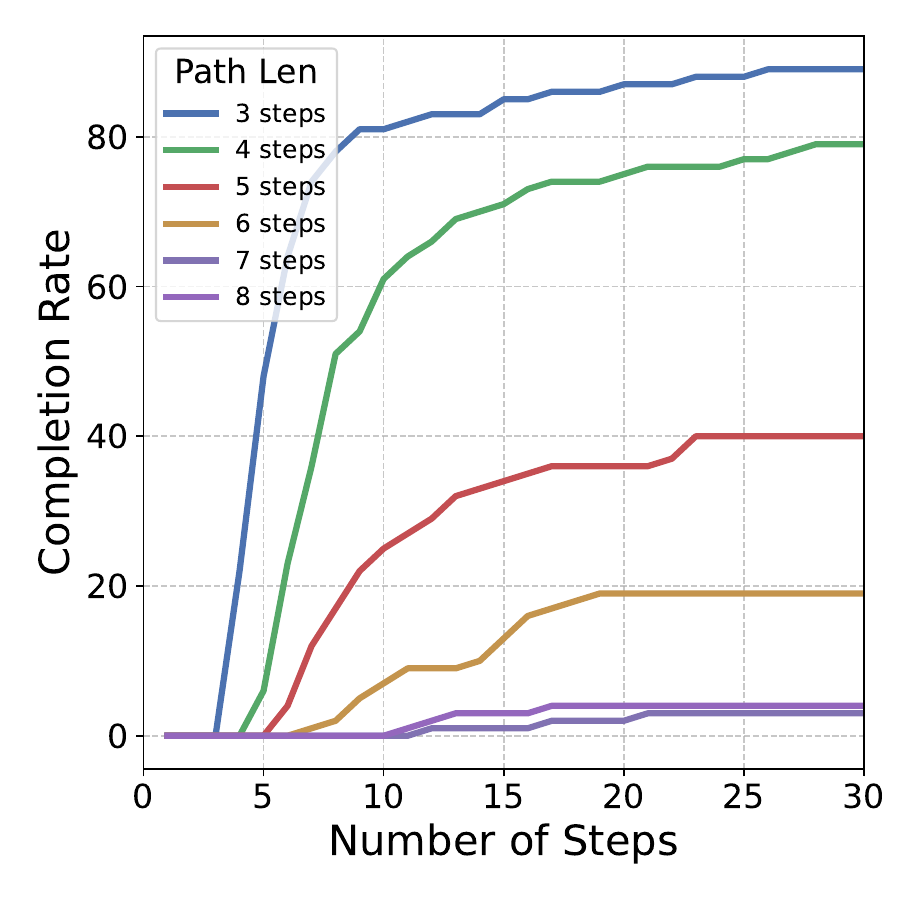}
        \caption{Completion rate}
        \label{fig:completion_rate}
    \end{subfigure}
    \caption{(a) In LLM-WikiRace, the agent navigates the Wikipedia hyperlink graph from a source page (e.g., \emph{Banana}) to a target page (e.g., \emph{Ferrari}). At each step, it receives the current page, the titles of its outgoing links, the target, and the history of pages visited, then selects one link to follow. (b) The completion rate --- the fraction of games solved --- plotted against the step budget across different shortest source-to-target path lengths for \textsc{Llama-3-70B-Instruct}, this experiment highlights how the number of steps per game and source to target shortest path length affect model performance.}
    \label{fig:wikirace_combined}
    \vspace{-2em}
\end{figure*}

To address this we introduce LLM-WikiRace, the first benchmark that requires \emph{interactive, multi-step planning} over the full Wikipedia hyperlink graph, asking the LLM to reason over extensive human knowledge. Our benchmark is based on the popular WikiRace game, where given a source and a target Wikipedia page, the player must traverse the Wikipedia hyperlink graph step by step, selecting links until it reaches the target or exhausts a fixed step budget, see \Cref{fig:WikiRace_illustration}. Success requires a model to retrieve world knowledge encoded in its weights and deploy it during multi-step planning — at each step, judging which outgoing link is most likely to advance toward the target. LLM-WikiRace is also useful for practitioners: by stripping away the domain-specific complications, tool-use quirks, and prompt sensitivities that dominate real-world agentic evaluations, it exposes flaws in a model's long-horizon planning ability. LLM-Wikirace is designed with cost and ease of use in mind, the benchmark presents two easy-to-use modes, one for models served by API and another for open-source models run locally. Our implementation runs multiple games in parallel cutting the benchmark run time in half, evaluating \textsc{Llama-3-70B-Instruct} in parallel on one split of the benchmark takes 3 hours compared to a synchronous implementation which takes 6 hours. All our code is open-source \footnote{\url{https://github.com/JuliuszZiomek/LLM-WikiRace-Benchmark}}. Our key contributions are as follows:

\begin{itemize}[noitemsep]

\item \textbf{LLM-WikiRace challenges frontier models on the hardest difficulty.}

State of the art frontier models — Gemini 3, GPT-5, Claude Opus 4.5 — succeed on fewer than 30\% of hard instances, leaving plenty of room for improvement. We provide a comprehensive study of over 20 models ranging from small open-source models (e.g. \textsc{Gemma-3-4B}) to large frontier models. 

\item \textbf{Models plan well, but replan badly.}

Trajectory-level analysis of model planning behaviours show that, surprisingly, despite adopting sensible high-level strategies, models fail in relatively naive ways - crucially they struggle to replan after failure becoming trapped in loops. 

\item \textbf{A measurable planning gap between reasoning and non-reasoning models} 

 Among open-source instruct-tuned models, success on LLM-WikiRace is well-predicted by a static world-knowledge probe. Reasoning models break this relationship: at similar knowledge levels, success rates differ by up to 20 percentage points. We label this the planning gap, which we hypothesize reflects the contribution of reasoning post-training to a model’s planning ability.
\newpage
\item \textbf{Post-training has limits.}

Finally, we show that post-training open-source models can improve their reasoning capabilities on the easy split of the benchmark by \textbf{$\sim$40\%}, but as the difficulty increases the gains from post-training decay and have no effect on the benchmark’s hardest difficulty.   

\end{itemize}

\section{What Type of Task Is WikiRace?}
\label{sec:What Type of Task Is WikiRace}
WikiRace is a task defined over the Wikipedia hyperlink graph in which an agent must navigate from a \emph{start page} $A$ to a \emph{target page} $B$ by selecting hyperlinks step by step. At each step, the agent observes the current page and its outgoing links, chooses a single link to follow, and transitions to a new page. The episode terminates when the target is reached or when a fixed step budget is exhausted. The objective is to reach the target in as few steps as possible. We illustrate this in \Cref{fig:WikiRace_illustration}.\looseness=-1

\begin{figure}

\begin{tcolorbox}[colback=black!4!white, colframe=blue, title=WikiRace Game Prompt]

\vspace{0.2cm}

\begin{small}
\textbf{System Prompt:}\\
You are a helpful assistant helping play the Wikipedia link game.\\

\textbf{User Prompt:}\\
You are playing a game where you start at Wikipedia page ``\{current\_page\}'' and want to reach page ``\{target\_page\}'' by clicking links.\\

So far, you have visited the following pages in order:\\
\{history\}\\

You see the following possible links from the current page:\\

0. \{Neighbor 0 Title\}\\
1. \{Neighbor 1 Title\}\\
$\vdots$\\
\{N\}. \{Neighbor N Title\}\\

Which link should you click to get closer to the target? Reply with the number of your choice (0 to \{max\_choice\_num\}).
\end{small}
\end{tcolorbox}
\caption{The LLM-WikiRace prompt, at each step of the game an LLM is prompted with the current page, the target page, a history of previously visited states and 50 possible next states.}
\label{fig:prompt}
\vspace{-2em}
\end{figure}

The task poses a challenging planning problem over a large, real-world knowledge graph, requiring agents to combine planning and world knowledge in a partially observable, interactive setting. At inference time, the agent has no access to global graph information such as shortest paths, node distances, or future link availability, and must act based only on locally observed information while leveraging world knowledge acquired during pretraining to judge which pages may be relevant. Each step presents many locally plausible actions whose long-term consequences are uncertain, and early choices can strongly constrain future options. As a result, poor initial decisions may move the agent far from the target, making effective replanning—rather than merely good initial planning—essential for avoiding failure. In particular, LLM-WikiRace probes the following capabilities of a reasoning system:\looseness=-1


\textbf{Long-horizon planning.}
Reaching the target often requires selecting intermediate pages that are not directly related to the goal, testing the ability to plan coherent action sequences rather than rely on myopic heuristics.\looseness=-1

\textbf{Semantic reasoning over open-domain knowledge.}
Effective navigation depends on reasoning about relationships between concepts encoded in Wikipedia's hyperlinks.\looseness=-1

\textbf{Operational use of world knowledge.}
Rather than answering questions, agents must translate world knowledge into concrete action choices that guide navigation.\looseness=-1

\textbf{Replanning and adaptive control.}
When an initial strategy proves ineffective, successful agents must revise their approach, avoid repeated states, and explore alternative paths under a limited action budget.\looseness=-1
\subsection{What the Benchmark Is Not Measuring}

LLM-WikiRace is not equivalent to shortest-path graph search. Agents are not given access to the global graph structure and each step consumes the limited step budget. As such, the model cannot just algorithmically perform standard graph search procedures (such as BFS).\looseness=-1

\section{How Does the LLM Play WikiRace?}\label{sec:Methodology}


We build the LLM-WikiRace hyperlink graph from a Wikipedia snapshot dated 23 June 2025 and retain only its largest strongly connected component, guaranteeing that every  node is reachable from every other node. The resulting graph contains 549,232 pages. Fixing this snapshot makes results stable and directly comparable across models; preprocessing details are in Appendix \ref{app:tech}.

To adapt WikiRace for LLMs and explore the interaction between internal world knowledge and planning, we make several changes to the standard WikiRace game. Firstly, we provide the LLM with a structured prompt that specifies the current page, the target page, the traversal history, and the set of outgoing links available at each step; see \Cref{fig:prompt}. Unlike the original game, we exclude the body of the current article from the LLM-WikiRace prompt. The agent has already reached this page and must choose from among outgoing links based on their names alone, this matches the information available to a player in the original game. Excluding the article body further requires the agent to ground its position in pre-trained world knowledge rather than in-context text. Successful navigation also demands planning several steps ahead, thus the agent must anticipate the contents of unvisited pages. Including the article body would also more than double prompt length, inflating cost, and would demand either a local Wikipedia snapshot of roughly 100~GB or a flood of calls against the rate-limited Wikipedia API. Appendix~\ref{app:page context} reports an ablation on this design choice.

Secondly, we split WikiRace into three varying difficulty levels. We define an easy, medium, and hard difficulty using the shortest path length between the source and target pages as a measure of difficulty; we investigated this design decision on \textsc{Llama-3-70B-Instruct}, as can be seen in \Cref{fig:completion_rate}, the number of successfully completed games (completion rate) monotonically decreases as the path length increases. 
The easy split comprises 200 pairs with shortest path lengths of 3 or 4, medium comprises 150 pairs with lengths of 5 or 6, and hard comprises 100 pairs with lengths of 7 or 8, with each split evenly balanced across its two path lengths.


Each game is capped at 30 steps. Our \textsc{Llama-3-70B-Instruct} experiment in \Cref{fig:completion_rate} shows completion rate plateauing well before this limit across all path lengths, giving models more than three times the headroom of the longest shortest path. We also cap the link list shown to the model at 50; 71\% of pages in our graph have fewer than 50 outgoing links. For the remaining pages we keep the 50 links with the shortest residual path length to the target and present them in random order; this shortest-path information lives only inside the environment and is never revealed to the model, so it serves to trim the prompt rather than guide the agent's reasoning. We report results with random link selection in Appendix~\ref{app:init}.\looseness=-1


\section{Evaluation}


\textbf{Models.} We evaluate a diverse set of models on our benchmark, spanning both open- and closed-source systems across a wide range of scales and modalities. Our closed-source evaluations include the GPT-5 family \cite{singh2025openai}, Gemini 2.0–3.1 \cite{comanici2025gemini}, Claude Sonnet 4.5 and Opus 4.5 \& 4.6 \cite{anthropic2024claude3}, and Grok 4.1-Fast. For open-source models, we evaluate DeepSeek R1 \cite{gao2023scaling}, Kimi K2 \cite{team2025kimi}, LLaMA~3 at scales ranging from 1B to 70B parameters \cite{grattafiori2024llama}, Gemma~3 models from 4B to 27B parameters \cite{team2024gemma}, Apertus 8B and 70B \cite{hernandezcano2025apertus}, Mistral 7B \cite{jiang2023mistral} and Ministral 8B \cite{mistral2024ministraux}, and Qwen~2.5-7B \cite{qwen2025qwen25}. We additionally include two diffusion-based language models, Dream-v0-Instruct 7B \cite{ye2025dream} and LLaDA-Instruct 8B \cite{nie2025large}. For all open source models we test the Instruct finetuned version.\looseness=-1

\textbf{Metrics.} We report several complementary metrics to characterise model performance. First, we measure the fraction of games successfully completed on each split, referred to as the \emph{success rate}. Second, to assess solution quality among successful runs, we compute the number of \emph{suboptimal steps}, defined as the excess number of steps taken beyond the shortest possible path. This metric captures how efficiently a model navigates once it succeeds, separating path quality from binary success. Finally, we report the average cost per game, computed over all evaluated episodes, including the total number of tokens generated and, for closed-source models, the corresponding monetary cost.

\begin{figure*}[t!]
    \centering
    \includegraphics[width=0.32\linewidth]{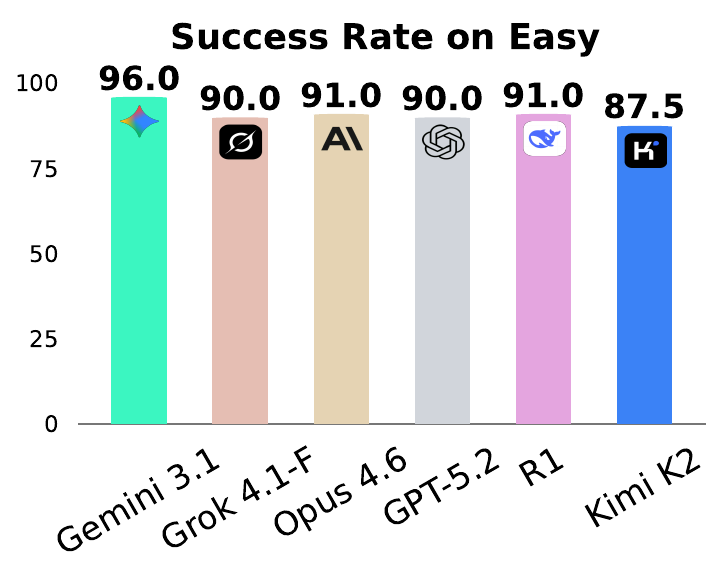}
    \includegraphics[width=0.32\linewidth]{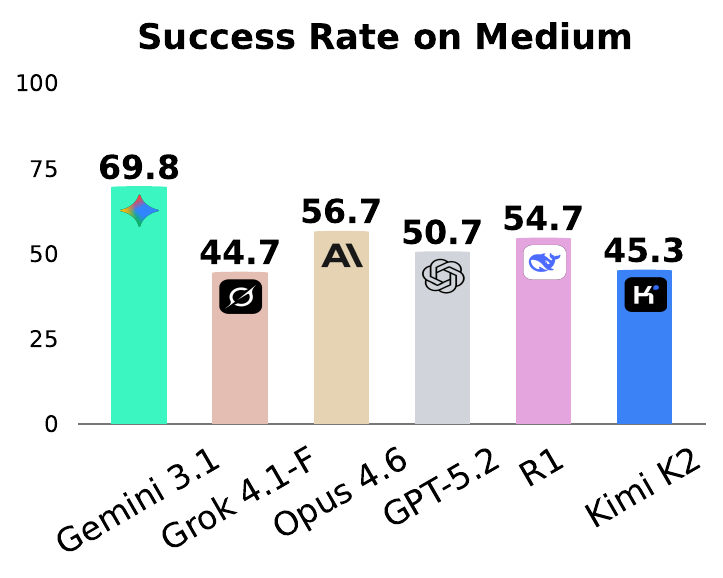}
    \includegraphics[width=0.32\linewidth]{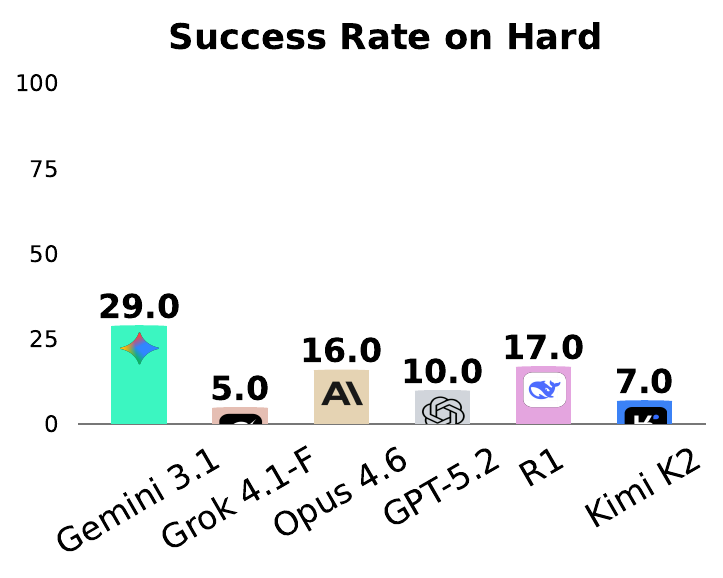}
    \caption{The Success Rates for the best performing models across all three difficulties of the LLM-WikiRace benchmark.}
    \label{fig:WikiRace_barplot}
\end{figure*}

\section{Results}

\begin{table*}[t]
\centering

\caption{Performance of different models across Easy, Medium, and Hard datasets. Success Rate is the percentage of games completed within the allowed 30 steps. Suboptimal Steps is the average number of suboptimal steps made in the successfully completed games. If no games were successfully completed for a given model and difficulty, we set Suboptimal Steps to "-". The best result is in \textbf{bold}, and the second best result is \underline{underlined} for each difficulty level's success rate.}
\label{tab:llm_WikiRace_results}

\begin{tabular}{@{}l@{\hspace{2pt}}l@{\hspace{2pt}}c@{\hspace{4pt}}c@{\hspace{4pt}}cc@{\hspace{4pt}}c@{\hspace{4pt}}cc@{\hspace{4pt}}c@{}}
\toprule
\textbf{Family} & \textbf{Model} & \multicolumn{3}{c}{\textbf{Success Rate (\%) $\uparrow$}} & \multicolumn{3}{c}{\textbf{Suboptimal Steps} $\downarrow$} & \multicolumn{2}{c}{\textbf{Usage Per Step}} \\
\cline{3-10}
 &  & \textbf{Easy} & \textbf{Medium} & \textbf{Hard} & \textbf{Easy} & \textbf{Medium} & \textbf{Hard} & \textbf{Tokens} & \textbf{Cost (cents)} \\
\midrule
\multirow{3}{*}{GPT} & GPT-5 Nano & 71.5 & 24.7 & 4.0 & 3.4 & 7.6 & 9.2 & 2170 & 0.07 \\
  & GPT-5 Mini & 85.5 & 46.0 & 11.0 & 2.2 & 3.9 & 4.6 & 874 & 0.11 \\
  & GPT-5 & 92.5 & 60.0 & 15.0 & 1.8 & 2.7 & 6.3 & 1826 & 1.51 \\
  & GPT-5.2 & 90.0 & 50.7 & 10.0 & 2.2 & 3.2 & 4.0 & 1008 & 0.85  \\
\midrule
\multirow{4}{*}{Gemini} & Gemini 2.0 Flash & 88.0 & 41.3 & 6.0 & 4.1 & 6.0 & 14.7 & 501 & 0.01 \\
  & Gemini 2.5 Flash & 91.0 & 53.0 & 10.2 & 2.7 & 4.5 & 8.1 & 547 & 0.05 \\
  & Gemini 2.5 & 91.0 & 56.7 & 15.2 & 2.1 & 3.5 & 6.9 & 527 & 0.19 \\
  & Gemini 3 & \underline{95.0} & \underline{66.0} & \underline{23.0} & 0.8 & 1.9 & 4.7 & 1848 & 1.76 \\
  & Gemini 3.1 & \textbf{96.0} & \textbf{69.8} & \textbf{29.0} & 1.5 & 3.0 & 5.8 & 1905 & 1.84 \\
\midrule
\multirow{2}{*}{Claude} & Sonnet 4.5 & 88.5 & 43.3 & 10.1 & 2.3 & 4.9 & 10.1 & 1242 & 1.31 \\
  & Opus 4.5 & 91.5 & 56.0 & 18.0 & 1.3 & 2.3 & 8.1 & 1906 & 3.73 \\
  & Opus 4.6 & 91.0 & 56.7 & 16.0 & 1.7 & 3.4 & 10.1 & 1354 & 2.38 \\
\midrule
\multirow{1}{*}{Grok} & Grok 4.1-Fast & 90.0 & 44.7 & 5.0 & 1.6 & 4.3 & 6.4 & 4458 & 0.21 \\ \hline 
\midrule
\multirow{1}{*}{DeepSeek} & R1 & 91.0 & 54.7 & 17.0 & 1.4 & 4.0 & 6.4 & 2598 & - \\
\midrule
\multirow{1}{*}{Kimi} & Kimi K2 & 87.5 & 45.3 & 7.0 & 2.3 & 5.0 & 4.9 & 8105 & - \\
\midrule
\multirow{4}{*}{LLaMA} & LLaMA 3 1B & 16.5 & 0.0 & 0.0 & 12.9 & - & - & 511 & - \\
  & LLaMA 3 3B & 47.0 & 3.3 & 0.0 & 11.4 & 13.2 & - & 783 & - \\
  & LLaMA 3 8B & 64.5 & 9.3 & 0.0 & 7.3 & 7.8 & - & 641 & - \\
  & LLaMA 3 70B & 84.5 & 39.3 & 7.0 & 2.5 & 4.2 & 5.1 & 651 & - \\
\midrule
\multirow{3}{*}{Gemma} & Gemma 3 4B & 48.0 & 2.7 & 0.0 & 5.2 & 11.5 & - & 555 & - \\
  & Gemma 3 12B & 72.5 & 22.7 & 1.0 & 3.4 & 6.8 & 4.0 & 651 & - \\
  & Gemma 3 27B & 80.0 & 30.0 & 0.0 & 2.8 & 6.4 & - & 684 & - \\
\midrule
\multirow{2}{*}{Apertus} & Apertus 8B & 42.0 & 4.0 & 0.0 & 10.2 & 16.5 & - & 805 & - \\
  & Apertus 70B & 65.0 & 10.7 & 0.0 & 8.2 & 13.9 & - & 1832 & - \\
\midrule
\multirow{2}{*}{Mistral} & Mistral 7B & 59.0 & 10.0 & 1.0 & 7.9 & 8.7 & 12.0 & 664 & - \\
  & Ministral 8B & 65.5 & 8.7 & 0.0 & 7.4 & 9.8 & - & 624 & - \\
\midrule
\multirow{1}{*}{Qwen} & Qwen 2.5-7B & 22.5 & 1.3 & 0.0 & 13.2 & 16.5 & - & 2597 & - \\
\midrule
\multirow{2}{*}{Diffusion} & Dreamv0-In. 7B & 53.0 & 3.3 & 1.0 & 8.7 & 11.2 & 12.0 & 1549 & - \\
  & LLaDA-In. 8B & 40.5 & 4.7 & 0.0 & 7.6 & 12.4 & - & 1669 & - \\
\bottomrule
\end{tabular}

\end{table*}

\begin{wrapfigure}{r}{0.45\textwidth}
\centering
\vspace{-1em}
\includegraphics[width=\linewidth]{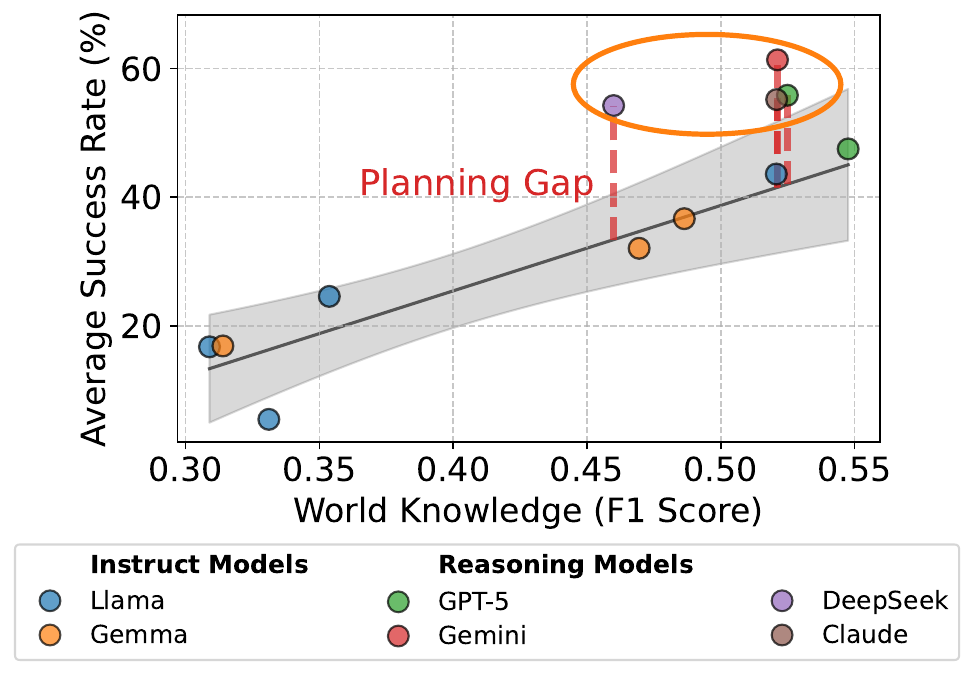}
\caption{LLM-WikiRace evaluates both internal world knowledge and planning, we identify a clear performance gap between the Reasoning models and Instruct tuned models despite both showing similar levels of world knowledge.}
\label{fig:planning gap plot}
\vspace{-1.5em}
\end{wrapfigure}

We present full results in Table~\ref{tab:llm_WikiRace_results} and summarise the performance of the strongest models in Figure~\ref{fig:WikiRace_barplot}. Performance is clearly differentiated by difficulty: leading models achieve over 90\% success on the easy split, approximately 50–70\% on medium, and below 30\% on hard. This confirms that the length of the shortest-path provides a meaningful stratification of task difficulty.\looseness=-1

Among the evaluated models, Gemini~3 and 3.1 perform best at all difficulty levels and are the only model to exceed a success rate of 20\% on the hard split. GPT-5 and Claude Opus~4.5/4.6 follow closely, with performance varying by difficulty. DeepSeek~R1 is competitive with these models despite being open source, typically trailing by a small margin, while Kimi~K2 ranks second among open source models with a more pronounced performance gap. Surprisingly, GPT-5.2 saw a notable decrease in performance on the medium and hard benchmarks in comparison to GPT-5. 

Despite strong performance on the easy split, both the medium and hard difficulties remain largely unsolved. Even the best-performing model completes fewer than 30\% of hard instances, indicating substantial headroom for improvement. Open source models from the LLaMA and Gemma families achieve near-zero success on the hard split, with the exception of \textsc{LLaMA-3-70B-Instruct}, which reaches a 7\% success rate. For these models, improving medium-difficulty performance—currently below 40\% across the board—represents a more attainable near-term target.

Overall, results across model families suggest that all three difficulty levels remain relevant at present, with each split providing a distinct and informative benchmark for models at different scales.\looseness=-1

\subsection{Does Graph Knowledge Affect Model Performance?}
\label{sec:World Knowledge}

To understand how the internal world knowledge of a model affects their performance on LLM-WikiRace we design a classification task to probe the model’s understanding of the Wikipedia hyperlink graph.  We construct a dataset consisting of $1k$ pairs of source and target nodes. Pairs are labeled as \emph{connected} if there exists a direct hyperlink from the source node to the target node and as \emph{unconnected} otherwise.  For each pair in this dataset we ask the model to predict if the nodes are directly connected or not, details of the dataset construction and model prompt for this task can be found in Appendix \ref{app:worldmodel}. We use the F1-Score of the model on this classification task as a metric for how well a model understands how pages are connected through the Wikipedia hyperlink graph.

In \Cref{fig:planning gap plot} we plot the F1-Score against the average success rate of each model across all LLM-WikiRace splits for a range of instruct and reasoning models. We fit a linear regression model on the instruct models and plot the 95\% confidence interval assuming independent gaussian noise. Among instruct-tuned models, the world knowledge probe is a reasonable predictor of success rate; however, reasoning models break this relationship. At similar levels of world knowledge, success rates differ by up to 20\% for reasoning models, lying outside the 95\% regression confidence interval. We label this phenomenon the Planning Gap, as reasoning models seem to be able to leverage their world knowledge more effectively than instruct tuned models. We hypothesize that this is due to their improved planning capabilities. 

In \Cref{fig:planning gap plot} one reasoning model, GPT-5-mini, is firmly within the confidence interval of the instruction-tuned models and does not exhibit a planning gap. We suspect that this is because the model is throttled to reduce its inference costs; however, since the exact implementation is unknown, we can only speculate. It is worth noting that the world knowledge probed by our classification task may differ from that required by LLM-WikiRace. Whereas classification entails a binary judgment about whether two specified pages are linked, the game demands open-ended retrieval of pages likely to be linked from a given source. This difference may affect the trends we found in our analysis. Despite these limitations, this experiment provides strong evidence that \emph{both} reasoning and world knowledge are necessary for success on LLM-WikiRace.

\subsection{Qualitative Analysis and Examples}
To better understand the behavioral differences that underlie model performance, we analyze reasoning traces from two of the strongest models that expose intermediate reasoning, Gemini~3 and Claude Opus~4.5 (GPT-5 does not provide explicit reasoning traces). For each model, we randomly sample three successful and three unsuccessful trajectories at each difficulty level, yielding 18 trajectories per model.\looseness=-1

\begin{figure*}[t]
  \centering
  \begin{qualexamplebox}{Example 1: Gemini 3 pro, \emph{Herstory} $\rightarrow$ \emph{Warth}}
    \small
        \annblock{orange}{Understanding target and current page}{The target page "Warth" most likely refers to a location (e.g., Warth, Vorarlberg in Austria; Warth, Thurgau in Switzerland) or a disambiguation page leading there. The current page "Herstory" is a sociological and linguistic concept regarding the perspective of history.}
        
        \annblock{blue}{Strategy}{To reach a specific geographical location from a sociological page, the best strategy is to move to the broadest possible topic that encompasses geography or nations.}
        \annblock{teal}{Consider options}{
        Among the options:\par
        \begin{itemize}\setlength\itemsep{2pt}
            \item \textbf{Link 25 (History)} is the parent concept of "Herstory". The page "History" is a vital, top-level article that links to broad categories like "Geography", "World", "Europe", and specific countries. This is the most efficient hub to transition from a specific feminist term to general locations.
            \item \textbf{Link 37 (Latin)} is a language. While it connects to European history, "History" is a more direct path to general knowledge and geography.
            \item Other links (Feminism, Neologism, etc.) keep the context too narrow.
        \end{itemize}}
        
        \annblock{red}{Commit to decision}{
        Therefore, "History" is the best link to widen the scope and eventually navigate to Austria, Switzerland, or lists of municipalities where "Warth" can be found. 25}

        \textbf{\textcolor{forestgreen}{[Solves the game succesfully by following path
        Herstory
        $\rightarrow$ History  $\rightarrow$ Geography $\rightarrow$ Berlin $\rightarrow$ Bregenz $\rightarrow$ Warth]}}
  \end{qualexamplebox}
  \caption{Raw response of Gemini 3 pro for the game instance \emph{Herstory} $\rightarrow$ \emph{Warth} (optimal path length $=4$), first step. The response highlights the model's knowledge of the start and target page as well as the overall strategy of moving towards a broad topic initially.}
  
  \label{fig:example_gemini1}
  \vspace{-1em}
  \end{figure*}

\textbf{Planning strategies and forward reasoning.}
Across inspected trajectories, both models frequently employ sensible high-level strategies. In particular, they often navigate from narrow topics toward broader, highly connected pages early in the episode, a behaviour reminiscent of common human WikiRace strategies. Gemini~3 explicitly frames this “hub-seeking” strategy in 72\% of inspected games, while Claude Opus~4.5 does so in 94\%. Representative examples of this behaviour are shown in Figure~\ref{fig:example_gemini1}. In addition, both models frequently produce forward-looking reasoning, speculating about which pages may be encountered in future steps and outlining tentative multi-step plans toward the target.

\textbf{Replanning failures and looping behaviour.}
Despite exhibiting forward planning, both models frequently revisit the same pages multiple times, entering loops from which they fail to escape. Gemini~3 encounters loops in 66\% of the inspected trajectories, and Claude Opus~4.5 in 61\%. In nearly all such cases, the models explicitly acknowledge the repeated state in their reasoning traces, indicating awareness of being stuck. However, awareness does not reliably translate into effective strategy revision. In many cases, the models continue alternating between neighbouring pages until the step limit is reached.
This failure mode is especially pronounced on the hard split. Gemini~3 encounters loops in 86\% of hard tasks, of which only 10\% are solved successfully; by contrast, all hard tasks without loops are solved. Claude Opus~4.5 exhibits a similar pattern, encountering loops in 92\% of hard tasks, with a success rate of only 1\% among those trajectories.

\begin{wrapfigure}{l}{0.45\textwidth}
    \vspace{-1em}
    \centering
    \includegraphics[width=0.9\linewidth]{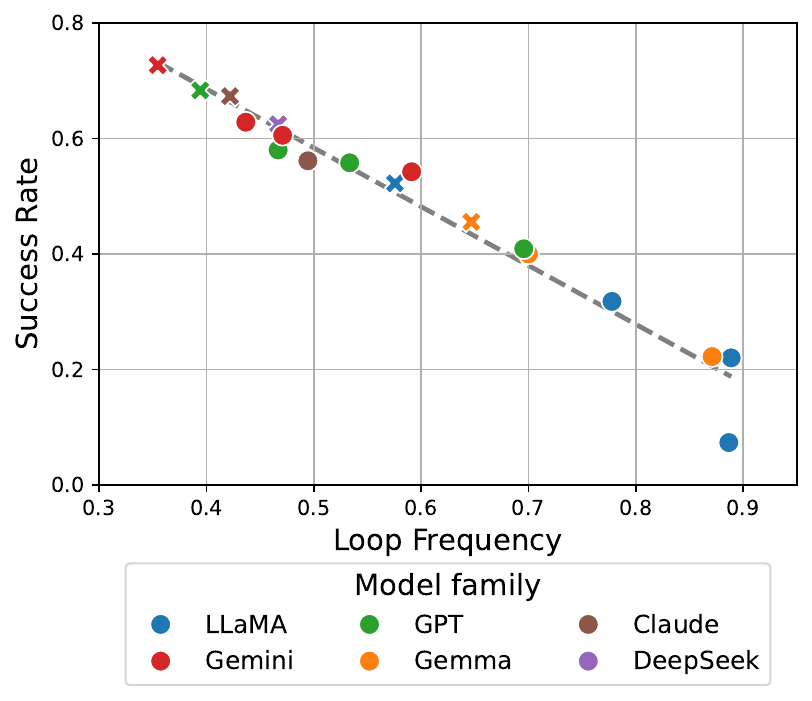}
    \caption{Success rate vs loop frequency for different models. We say that a game contains a loop if model visited any page more than once. Each model family is displayed with different colors on the scatter plot and the best model in each family is shown with an X mark. Dashed gray line shows linear regression fitted to points.}
    \label{fig:looping}
    \vspace{-2em}
\end{wrapfigure}

In Figure \ref{fig:looping} we plot the relationship between looping frequency and success rate for different model families. We see a strong and consistent linear trend, showing a clear negative correlation between the two quantities (regression coefficient $-1.02$, with 95\% CI $(-1.13, -0.91)$). This provides strong quantitative evidence for the general replanning failure trend we observed during qualitative analysis.

\textbf{Illustrative examples.}
Representative trajectories illustrating both successful and unsuccessful reasoning traces are provided in the Appendices \ref{ap:geminitraces}, \ref{ap:opustraces} and \ref{ap:geminiopustraces}, and are consistent with the aggregate patterns described above.

\textbf{Summary.}
Together, these qualitative findings show that frontier models can plan, reason about semantics, and monitor their own behaviour, but struggle with replanning—adjusting their strategy after observing that an existing approach is ineffective. Importantly, these replanning failures persist among the strongest models, highlighting remaining limitations that are not captured by aggregate success rates alone.

\subsection{Does Post-Training Help?}

\begin{wraptable}{r}{0.48\textwidth}
\centering
\caption{Success rate across difficulty levels for the base model and DAPO-finetuned checkpoints. Number after $FT-$ is the number of training steps model was finetuned for.}
\label{tab:finetuning_results}
\begin{tabular}{lccc}
\toprule
Model & Easy & Medium & Hard \\
\midrule
Base   & 22.5 & 1.3 & 0.0 \\
FT-50  & 24.0 & 1.3 & 0.0 \\
FT-100 & 41.0 & 4.0 & 0.0 \\
FT-300 & \textbf{67.5} & \textbf{4.6} & 0.0 \\
\bottomrule
\end{tabular}
\end{wraptable}

We conduct a preliminary study to investigate if RL post-training can improvement the ability of models to reason on the benchmark. We construct a dataset of 1{,}000 source and target page pairs with shortest path length ranging from 2 to 6 steps; and ensure there is overlap with pairs in the easy or medium difficulty splits. Each training example consists of a game prompt (see \Cref{fig:prompt}) instantiated at the source page and a set of candidate links. We use the DAPO algorithm \cite{shao2024deepseekmath,guo2025deepseek, yu2025dapo}  to train a Qwen-2.5-7B base model on this dataset. The model receives a binary reward signal, a reward of one is given if it reasons and selects the link with the shortest path length to the target page and a reward of zero otherwise. We train for up to 300 steps stopping when the validation reward plateaus, and evaluate checkpoints at steps 50, 100, and 300 on the easy, medium, and hard splits. For full details, including training hyperparameters, see Appendix \ref{app:fine-tune}.  

As shown in Table~\ref{tab:finetuning_results}, fine-tuning leads to substantial improvements on the easy split, with success rates increasing from 22.5\% for the base model to 67.5\% after 300 fine-tuning steps. Performance on the medium split improves more modestly, from 1.3\% to 4.6\%, while performance on the hard split remains at 0 across all evaluated checkpoints. Taken together, these results indicate that, while DAPO-based fine-tuning can meaningfully improve performance on simpler instances, it does not, in this preliminary setting, address the challenges posed by harder LLM-WikiRace tasks.

\section{Related Work}

Many LLM benchmarks require planning either as a primary objective or as a subtask for success \cite{li2025planet}. Prior work has focused on classical planning settings, including benchmarks such as PlanBench \cite{valmeekam2023planbench} and structured games like Blockworld \cite{gupta1992complexity}, Sudoku \cite{seely2025sudoku}, and related environments \cite{wu2023smartplay, huang2024far, dagan2024plancraft, duan2024gtbench}. Analyses of these tasks \cite{valmeekam2024llms, valmeekam2025systematic, kambhampati2024llms} report mixed planning performance and question whether LLMs perform genuine planning or rely on pattern retrieval. Planning is also central to multimodal, embodied benchmarks \cite{shridhar2020alfred, shridhar2020alfworld, chen2023egoplan, su2024actplan, chang2024partnr, yang2025embodiedbench, yin2024safeagentbench}. In contrast, LLM-WikiRace isolates long-horizon planning and world knowledge within a single textual modality.

\textbf{Realistic and open-domain planning.}
Recent benchmarks have moved toward more realistic settings by embedding planning within naturalistic tasks, including travel itinerary construction \cite{xie2024travelplanner}, personal scheduling \cite{zheng2024natural}, agentic frameworks where planning is required to solve coding problems \cite{liu2023agentbench, jimenez2023swe, deng2025swe, xu2024theagentcompany}, and web-based assistant environments \cite{yoran2024assistantbench, zhou2023webarena, koh2024visualwebarena, deng2023mind2web}. While these benchmarks introduce real-world constraints and complexity, they are typically limited to specific domains. In contrast, LLM-WikiRace evaluates planning over a large, open-domain knowledge graph, requiring models to integrate long-horizon planning with broad, general world knowledge across a wide range of topics.

\textbf{Game-based benchmarks.}
BALROG \cite{paglieri2024balrog} evaluates planning and reasoning across a collection of game environments with varying horizons. LLM-WikiRace complements this line of work by emphasizing open-domain knowledge: success depends on understanding and exploiting relationships across a wide range of real-world concepts encoded in the Wikipedia hyperlink graph.

Concurrent work \cite{margiotta2025evaluating} uses the WikiRace game to contrast LLM and human reasoning capacities and \cite{ebrahimi2025textual} explores planning algorithms on a similar setting. Our work is an LLM focused benchmark and conducts a broader analysis of available models. Furthermore, we specify our benchmark independent of a human notion of difficulty and do not allow models to follow an algorithmic framework forcing the LLM to reason unaided on the graph.\looseness=-1

\section{Future Work and Conclusion}

LLM-WikiRace is built on the Wikipedia hyperlink graph, yielding a diverse set of verifiable tasks that probe planning and reasoning over varying horizon lengths. Our analysis exposes several promising directions for future work. In particular, our initial fine-tuning experiments reduce the game to a one-step reasoning problem during training, which could naturally be  extended to multi-step reinforcement learning approaches. More broadly, understanding how planning frequency \cite{paglieri2025learning}, retrieval-augmented generation \cite{gao2023retrieval}, and other agentic frameworks influence long-horizon reasoning remains an open and fertile area of study.

LLM-WikiRace evaluates an LLM’s ability to integrate world knowledge into planning and decision-making. While frontier models perform well on the easiest split, none achieve a success rate above $30\%$ on the hardest split. We highlighted how strong reasoning capabilities differentiate the best models despite showing similar levels of knowledge about the underlying graph the game is based on. Even these models exhibit systematic planning failures on difficult tasks, including looping behavior, reluctance to adapt initial strategy, and insufficient exploration. Overall, LLM-WikiRace is easy for models to understand yet difficult to master, exposing persistent limitations in long-horizon reasoning and planning when models must act on their world knowledge.

\bibliography{ref}
\bibliographystyle{plainnat}

\newpage
\onecolumn

\appendix

\let\addcontentsline\OrigAddContentsLine
\addcontentsline{toc}{section}{Appendix} 
\part{Appendix} 
\parttoc 


\newpage
\section{Technical details of the implementation}
\label{app:tech}

\begin{algorithm}[H]
\caption{WikiRace Game Procedure}
\label{alg:WikiRace}
\begin{algorithmic}[1]
\STATE \textbf{Input:} Source page $s$, target page $g$, maximum steps $T$, maximum links $N$
\STATE Initialize current page $p \gets s$
\FOR{$t = 1$ \textbf{to} $T$}
    \STATE Construct a prompt containing the current page $p$, target page $g$, traversal history, and outgoing links
    \STATE Retrieve outgoing pages $\mathcal{N}(p)$ from the WikiGraph
    \IF{$|\mathcal{N}(p)| > N$}
        \STATE Select up to $N$ outgoing links with the shortest precomputed distance to $g$
    \ENDIF
    \STATE Prompt the LLM to select one link
    \STATE Update $p \gets$ selected link
    \IF{$p = g$}
        \STATE \textbf{Success}
        \STATE \textbf{break}
    \ENDIF
\ENDFOR
\STATE \textbf{Failure}
\end{algorithmic}
\end{algorithm}

\paragraph{WikiGraph Structure.}
We represent the Wikipedia hyperlink graph as a sparse, directed graph in compressed sparse row (CSR) format. Each node corresponds to a Wikipedia page (mapped from page ID to title), and each directed edge represents an outgoing hyperlink. Formally, the graph is encoded as an adjacency matrix $A \in \{0,1\}^{N \times N}$, stored using two arrays:
(i) \texttt{row\_ptr} of length $N{+}1$, and
(ii) \texttt{col\_indices} of length $\mathrm{nnz} = \lvert\{(i,j) : A_{ij}=1\}\rvert$.

For each node $i \in \{0,\dots,N{-}1\}$, the slice
\[
\texttt{col\_indices[row\_ptr[i] : row\_ptr[i+1]]}
\]
returns the indices of all out-neighbours of $i$, corresponding to the targets of outgoing hyperlinks from page $i$. This representation enables $O(1)$ access to a node’s neighbour list boundaries and requires $O(N + \mathrm{nnz})$ memory, with good cache locality for row-wise traversal.

\paragraph{Shortest-Path Computation.}
To compute shortest-path distances between pages, we use the SciPy \cite{virtanen2020scipy} function \texttt{scipy.sparse.csgraph.shortest\_path} over the sparse WikiGraph. Shortest-path distances from each target page are precomputed once and cached prior to gameplay. These distances are used exclusively by the environment to determine task difficulty and to filter outgoing links, and are never exposed to the model during inference.

Finally, a detailed algorithm of how the LLM-WikiRace game is played can be found in \Cref{alg:WikiRace}.

\textbf{Compute Resources.} For all experiments the open source models with $\leq 70B$ parameters were run locally on 2 H200 GPUs. Larger models were run through the open router API.

\newpage
\section{How was the difficulty split decided?} \label{app:init}

\begin{figure}[h]
    \centering
    \begin{subfigure}{0.32\linewidth}
        \centering
        \includegraphics[width=\linewidth]{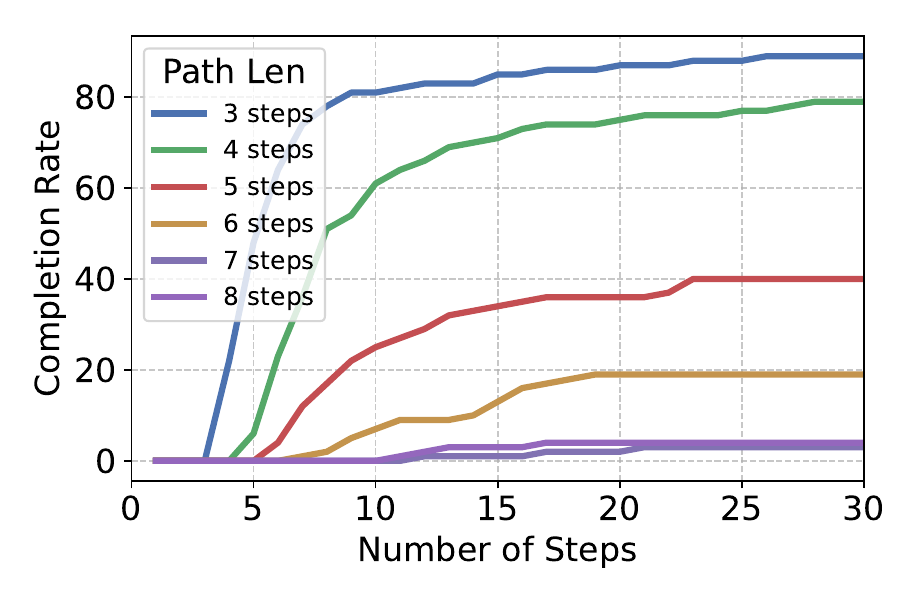}
        \caption{Completion rate}
    \end{subfigure}\hfill
    \begin{subfigure}{0.32\linewidth}
        \centering
        \includegraphics[width=\linewidth]{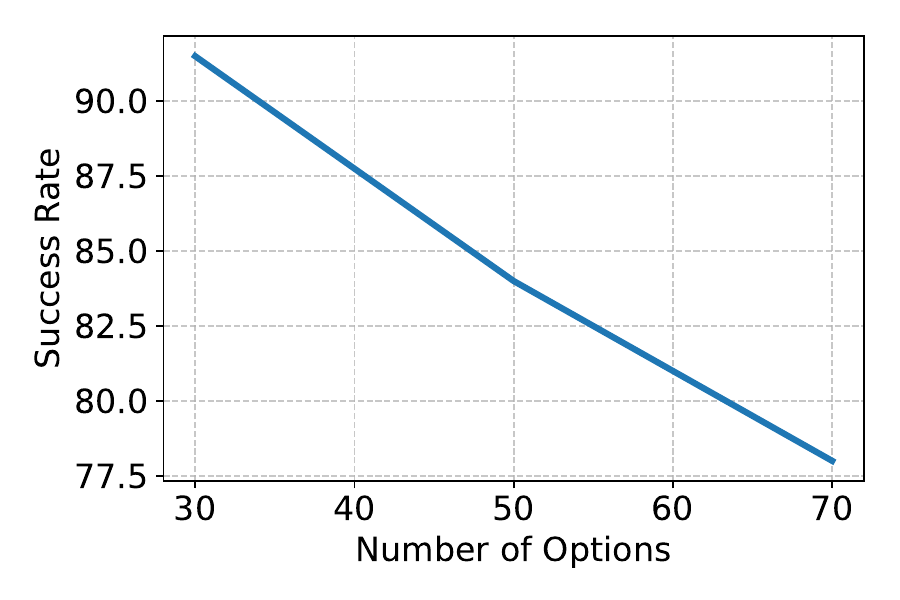}
        \caption{Options vs. success}
    \end{subfigure}\hfill
    \begin{subfigure}{0.32\linewidth}
        \centering
        \includegraphics[width=\linewidth]{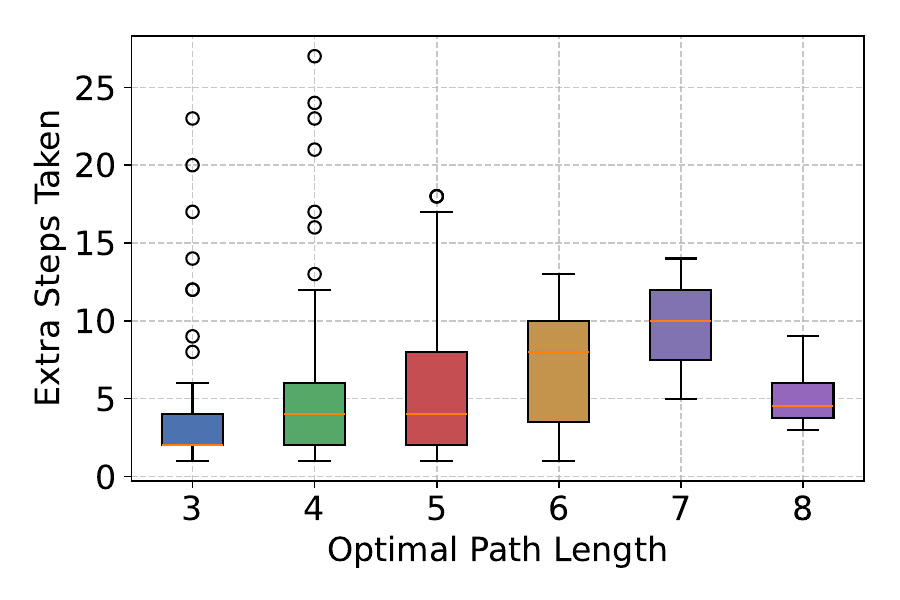}
        \caption{Extra steps}
    \end{subfigure}
    \caption{\textbf{We create easy, medium, and hard difficulty splits using the length of the shortest path between the start and target concept to determine difficulty.} \textbf{a)} The completion rate is plotted against the number of game steps for \textsc{Llama-3.3-70B-Instruct}, over a variety different optimal path lengths. Across different path lengths most models consistently pass or fail by step 30. \textbf{b)} We analyse the performance of the easy split, when the model is provided with $\{30,50,70\}$ next node options, we settle on $50$ options at each step to ensure the easy split is not saturated.  \textbf{c)} The extra steps taken for successful runs increases as the optimal path length increases, we hypothesize this is because more sub-optimal moves accumulate over longer tasks and require more steps to correct.}
    \label{fig:analysis of llama-3-70B}
\end{figure}

To better understand how large language models approach the task, we 
analyse the performance of \textsc{Llama-3.3-70B-Instruct} across a range of experimental settings. In \Cref{fig:analysis of llama-3-70B} a), we plot the average completion rate, whether the LLMs successfully reaches the target page within $T$ steps, over 50 games as a function of the maximum number of allowed steps T, stratified by the minimum path length, $O$, between the source and target pages. We observe a clear relationship between task difficulty—as measured by the shortest-path distance, and model performance: completion rates decrease as the minimum path length increases.

Increasing the number of allowed steps initially improves performance, as it enables the model to recover from early suboptimal decisions. However, the benefits of additional planning steps diminish beyond a certain point, and performance eventually plateauing despite increasing compute. In \Cref{fig:analysis of llama-3-70B} b), we analyse how the number of options presented to the LLM affects performance at step 30 on paths with lengths 3 \& 4. We found, as one might expect, that more options makes the task harder. Based off this experiment we set the number of options to be $50$ at each timestep, we did not pick $70$ to reduce the number of tokens in each prompt and further reduce the cost of running LLM-WikiRace.

In \Cref{fig:analysis of llama-3-70B} c), we further analyze the number of steps taken in excess of the minimum path length. This metric captures the extent to which the model deviates from optimal play. We find that suboptimality increases with the optimal path length, suggesting that small errors compound over longer trajectories. Based on these exploratory results, we fix the maximum number of allowed steps to $T=30$ for all subsequent experiments.

Based on this initial exploration, we formulate three LLM-WikiRace difficulty splits: \textsc{easy}, \textsc{medium}, and \textsc{hard}. These splits are created by sampling source pages at random from the graph and selecting a target pages at minimum $O$ pages away. For the easy split we select 200 nodes with the shortest possible path equally split between 3 and 4 pages, the medium baseline consists of 150 nodes with an equal split between 5 and 6 pages, finally the hard baseline consists of 100 pairs equally split between lengths 7 and 8. For the harder baselines models often require more steps, to reduce the cost of running the baseline we reduce the number of points to be evaluated as the difficulty of the splits increases.

\newpage
\section{Ablation of Action Space Trimming Method} \label{ap:link_prunning}

In this section we explore how our link pruning approach affects the performance of a range of models on LLM-WikiRace. In \Cref{tab:llm_WikiRace_results} we show results for the setting where the top 50 links ordered by the length of the shortest path from that page to the target page are presented to the model in a random order. As discussed in \Cref{sec:Methodology} around 70\% of the nodes in the preprocessed hyperlink graph have less than 50 outgoing links, leaving just 30\% of nodes in the graph effected by this design decision. To provide a thorough analysis of our design choices we also investigate how selecting links randomly affects a range of open source and closed source models. The results shown in \Cref{tab:closed source link ablation} and \Cref{tab:open source link ablation} compare the random link selection success rate with the top-50 link selection results reported in \Cref{tab:llm_WikiRace_results}.

The majority of models show little to no difference in performance, as measured by success rate, when switching from top-50 shortest path link selection to random link selection. The exception to this is Gemini 2.5-Flash which sees a huge drop in performance of 10\% and 20\% on the easy and medium difficult splits respectively. This large swing in performance is not observed in the open source model results, which shows comparable performance between the random and top-50 link selection settings.\looseness=-1

\begin{table}[h]
    \centering
    \begin{tabular}{cc|ccc}
     Model  & Link Selection  & Easy & Medium & Hard \\
     \midrule
     GPT-5 Mini    & Top-50  & 85.5 & 46.0 & 11.0 \\
     GPT-5 Mini & Top-1 + Random & 86.5 &  46.98  & 7.00 \\
     \midrule
     Gemini 2.5-Flash & Top-50 & 91.0 & 53.00 & 10.2 \\
     Gemini 2.5-Flash & Top-1 + Random & 82.0 & 33.33 & 3.00 \\
    \end{tabular}
    \caption{Closed source success rate (\%) under different link selection strategies}
    \label{tab:closed source link ablation}
\end{table}

\begin{table}[h]
\centering
\begin{tabular}{llccc}
\toprule
Model & Link Selection & Easy & Medium & Hard \\
\midrule
Apertus 70B & Top-50            & 65.0 & 10.7 & 0.0 \\
Apertus 70B & Top-1 + Random    & 65.0 & 12.0 & 0.0 \\
\midrule
Apertus 8B  & Top-50            & 42.0 & 4.0  & 0.0 \\
Apertus 8B  & Top-1 + Random    & 41.5 & 4.7  & 0.0 \\
\midrule
Gemma 3 12B & Top-50            & 72.5 & 22.7 & 1.0 \\
Gemma 3 12B & Top-1 + Random    & 70.5 & 20.7 & 1.0 \\
\midrule
Gemma 3 27B & Top-50            & 80.0 & 30.0 & 0.0 \\
Gemma 3 27B & Top-1 + Random    & 80.5 & 30.0 & 0.0 \\
\midrule
Gemma 3 4B  & Top-50            & 48.0 & 2.7  & 0.0 \\
Gemma 3 4B  & Top-1 + Random    & 45.0 & 4.0  & 0.0 \\
\midrule
LLaMA 3 1B  & Top-50            & 16.5 & 0.0  & 0.0 \\
LLaMA 3 1B  & Top-1 + Random    & 16.5 & 0.0  & 0.0 \\
\midrule
LLaMA 3 3B  & Top-50            & 47.0 & 3.3  & 0.0 \\
LLaMA 3 3B  & Top-1 + Random    & 51.5 & 6.7  & 0.0 \\
\midrule
LLaMA 3 70B & Top-50            & 84.5 & 39.3 & 7.0 \\
LLaMA 3 70B & Top-1 + Random    & 84.5 & 39.3 & 7.0 \\
\midrule
LLaMA 3 8B  & Top-50            & 64.5 & 9.3  & 0.0 \\
LLaMA 3 8B  & Top-1 + Random    & 67.5 & 14.0 & 1.0 \\
\bottomrule
\end{tabular}
\caption{Open Source success rate (\%) under different link selection strategies}
\label{tab:open source link ablation}
\end{table}

\newpage
\section{Ablation on Page Context}
\label{app:page context}

As part of the LLM-WikiRace prompt, shown in \Cref{fig:prompt}, we provide the model the current page, target page, and available links; we exclude the article body for the current page. We do this for a variety of reasons, practical and principled, covered in \Cref{sec:Methodology}. To thoroughly ablate the design of LLM-WikiRace we briefly conduct an ablation experiment of this design decision over a range of open source models. We generate a new prompt, shown in \Cref{fig:context prompt}, and use the Wikipedia API to add a summary of the \{current\_page\} to the prompt. The results are shown in \Cref{tab:context results}.

Including a summary of the current page in the prompt does not consistently help models. The Apertus-8B and two Gemma-3 models show improvements of $\sim$5\% on the easy split but no significant performance improvements on the medium and hard difficulties. For the Llama-3 models and the Qwen-2.5-7B-Instruct model, we observe that including this additional context reduces the model performance on the easy split, in some cases by up to 8\%, and dramatically hurts the model performance on the medium difficulty.

\begin{figure}[H]
\begin{tcolorbox}[colback=black!4!white, colframe=blue, title=WikiRace Context Game Prompt]

\vspace{0.2cm}

\begin{small}
\textbf{System Prompt:}\\
You are a helpful assistant helping play the Wikipedia link game.\\

\textbf{User Prompt:}\\
You are playing a game where you start at Wikipedia page ``\{current\_page\}'' and want to reach page ``\{target\_page\}'' by clicking links.\\

Context for the current page (``\{current\_title\}``):\\
``\{current\_context\}``\\

So far, you have visited the following pages in order:\\
\{history\}\\

You see the following possible links from the current page:\\

0. \{Neighbor 0 Title\}\\
1. \{Neighbor 1 Title\}\\
$\vdots$\\
\{N\}. \{Neighbor N Title\}\\

Which link should you click to get closer to the target? Reply with the number of your choice (0 to \{max\_choice\_num\}).
\end{small}
\end{tcolorbox}
\caption{The LLM-WikiRace prompt with current page context added.}
\label{fig:context prompt}
\vspace{-2em}
\end{figure}

\begin{table}[H]
    \centering
    \begin{tabular}{llccc}
\hline
Model & Page Context & Easy & Medium & Hard \\
\hline
Apertus 8B & False & 41.5 & 4.7 & 0.0 \\
Apertus 8B & True  & 46.5 & 6.7 & 0.0 \\
\hline
Gemma 3 12B & False & 70.5 & 20.7 & 1.0 \\
Gemma 3 12B & True  & 75.5 & 20.0 & 0.0 \\
\hline
Gemma 3 27B & False & 80.5 & 30.0 & 0.0 \\
Gemma 3 27B & True  & 82.0 & 29.3 & 0.0 \\
\hline
LLaMA 3 3B & False & 51.5 & 6.7 & 0.0 \\
LLaMA 3 3B & True  & 31.0 & 0.7 & 0.0 \\
\hline
LLaMA 3 8B & False & 67.5 & 14.0 & 1.0 \\
LLaMA 3 8B & True  & 63.5 & 5.3 & 1.0 \\
\hline
Qwen/Qwen2.5-7B-Instruct & False & 49.0 & 4.0 & 0.0 \\
Qwen/Qwen2.5-7B-Instruct & True  & 41.0 & 3.3 & 0.0 \\
\hline
\end{tabular}
\caption{Results for the context prompt vs. no-context prompt ablation experiment of LLM WikiRace}
\label{tab:context results}
\end{table}

\newpage
\section{World Model Knowledge Testing Details} \label{app:worldmodel}

\begin{figure}[H]

\begin{tcolorbox}[colback=black!4!white, colframe=blue, title=WikiRace World Model Experiment Prompt]

\vspace{0.2cm}

\begin{small}

\textbf{User Prompt:}\\
We are playing a game where you navigate Wikipedia from a starting page to a target page solely by clicking links on each page. 

Can you tell us if $\{$source\_page$\}$ contains a link to $\{$target\_page$\}$? Reply with either 'yes' or 'no' in the following format:\texttt{\textbackslash boxed$\{\}$}.
\end{small}
\end{tcolorbox}
\caption{A schematic of the prompt provided to the LLM, when asking it to select the link to follow}
\label{fig:world model test prompt}
\end{figure}

We now provide further details on the world knowledge experiment. To disentangle planning and world-knowledge we construct a classification task to probe the model's understanding of the underlying Wiki-graph. We ask the model to differentiate between pairs of connected and unconnected nodes. To better characterize different types of non-connectivity, we include several classes of unconnected pairs. Specifically, we sample target nodes that are $i$ steps away from the source node for $i \in \{2,3,4\}$, with $200$ samples included for each distance, and $200$ samples of directly connected nodes. In addition, we include a \emph{reversed connection} category, designed to be particularly challenging, in which the target node links to the source node but not visa-versa. For each source–target pair, we prompt the model using the prompt in \Cref{fig:world model test prompt} to predict whether the two nodes are connected, and parse the resulting binary yes/no response in a \texttt{\textbackslash boxed$\{\}$} format. In the smaller models if a response is not parsed we disregard the sample when calculating the world knowledge F1-score.

\newpage
\section{Post-training Experiment}
\label{app:fine-tune}

\begin{table}[h]
    \centering
    \small
    \begin{tabular}{ll}
        \toprule
        \textbf{Hyperparameter} & \textbf{Value} \\
        \midrule
        \multicolumn{2}{l}{\textit{Model \& Algorithm}} \\
        Base model              & Qwen2.5-7B \\
        RL algorithm            & DAPO \\
        KL penalty              & low-variance KL \\
        \midrule
        \multicolumn{2}{l}{\textit{Optimization}} \\
        Learning rate           & $1 \times 10^{-6}$ \\
        LR schedule             & constant \\
        LR warmup ratio         & 0.0 \\
        Gradient clipping       & 1.0 \\
        Total training steps    & 300 \\
        Train batch size        & 64 \\
        PPO mini-batch size     & 16 \\
        PPO epochs              & 1 \\
        \midrule
        \multicolumn{2}{l}{\textit{DAPO Objective}} \\
        Clip ratio (low)        & 0.2 \\
        Clip ratio (high)       & 0.28 \\
        KL loss coefficient     & $1 \times 10^{-3}$ \\
        Entropy coefficient     & 0.0 \\
        Loss aggregation        & token-mean \\
        Discount $\gamma$       & 1.0 \\
        GAE $\lambda$           & 1.0 \\
        \midrule
        \multicolumn{2}{l}{\textit{Rollout / Sampling}} \\
        Rollouts per prompt ($n$) & 16 \\
        Validation rollouts     & 8 \\
        Sampling temperature    & 1.0 \\
        Top-$p$ / Top-$k$       & 1.0 / $-1$ \\
        Max prompt length       & 1024 \\
        Max response length     & 1024 \\
        \midrule
        \multicolumn{2}{l}{\textit{Filtering}} \\
        Rejection sampling      & enabled \\
        Rejection multiplier    & 3.0 \\
        Max generation batches  & 5 \\
        \bottomrule
    \end{tabular}
    \caption{Training hyperparameters for DAPO fine-tuning experiment on LLM-WikiRace.}
    \label{tab:hyperparams}
\end{table}

\begin{figure}[h]
    \centering
    \includegraphics[width=\linewidth]{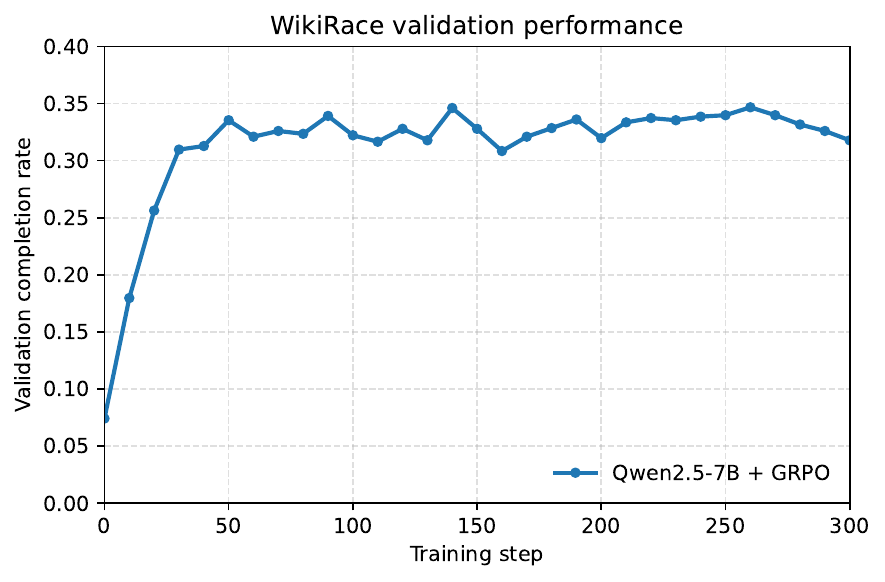}
    \caption{Validation completion rate on LLM-WikiRace throughout DAPO fine-tuning of Qwen2.5-7B. The model improves rapidly during the first $\sim$30 steps, climbing from a $7.4\%$ zero-shot baseline to over $30\%$, before plateauing around $33$--$35\%$ for the remainder of training.}
    \label{fig:wikirace_val_curve}
\end{figure}

\newpage
\section{Comparison with Human Performance}

To provide a coarse reference for human gameplay, we use publicly released statistics from The WikiGame Daily\footnote{\url{https://www.thewikigamedaily.com}}, collected from daily WikiRace games announced on the platform’s X account\footnote{\url{https://x.com/dailywikigame}}. We identify the corresponding start–target page pairs in our Wikipedia graph and refer to this set as the \textit{Human Gameplay Corpus}. We then evaluate the three strongest models—Gemini~3, GPT-5, and Claude Opus~4.5—on this corpus.

\begin{figure}[H]
    \centering
    \includegraphics[width=0.7\linewidth]{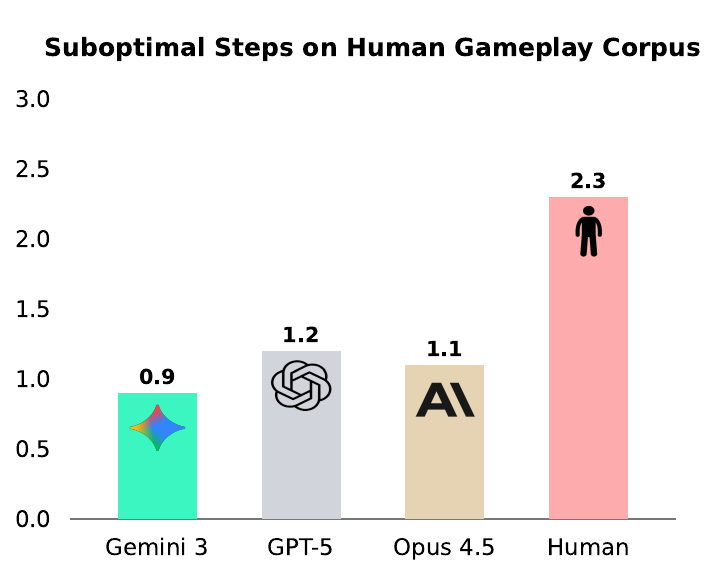}
    \caption{Comparison of suboptimal steps made by the most powerful models and by Human players, across the Human Gameplay Corpus games. The success rate on this corpus for was 98.5\% for humans and 100\% for all tested models.}
    \label{fig:human}
\end{figure}

All three models achieve a success rate of 100\% on the Human Gameplay Corpus, while the reported human success rate is 98.5\%. This indicates that the corpus is easier than the easiest split of our benchmark. As success rates are saturated, we instead compare the number of suboptimal steps taken by models and human players. Results are shown in Figure~\ref{fig:human}. On average, human players take at least one more suboptimal step than any of the evaluated models.

We emphasise that this corpus should be interpreted as a coarse reference rather than a precise estimate of human performance, due to heterogeneous player skill and potential graph reconstruction effects, human players are presented with all the links on a specific page whilst we limit the number of links in LLM-WikiRace. Nevertheless, the evaluated models consistently exceed this reference in terms of path optimality.\looseness=-1

\newpage
\section{Additional Analysis of Loops and Recovery}

To disentangle navigational precision from agentic adaptability / replanning abilities, we analyze the agent's behaviour regarding state revisitation. We define a \textit{loop} as any instance where an agent visits a specific page more than once within a single trajectory. We utilize three metrics to capture different aspects of this behaviour: \begin{itemize} 
\item \textbf{Loop Frequency (Precision):} The proportion of tasks where a loop occurred. This measures the agent's ability to navigate the graph linearly without errors. 
\item \textbf{Recovery Rate (Adaptability):} Conditioned on a loop occurring, this measures the proportion of tasks where the agent eventually solves the task. This serves as a proxy for the agent's capacity to recognize it is in a cycle and adapt its strategy successfully to escape and solve the task. 
\item \textbf{Average Maximum Visitation (Stagnation):} The average maximum number of times a single page is visited. High values here indicate "blind" looping, where the agent fails to realize it is repeating actions.
\end{itemize}

This correlation persists in the Medium split but shifts along the diagonal as task complexity increases. In the Hard split, all models are clustered in the bottom-right corner. Here, the difficulty of the tasks forces nearly all models into frequent looping behaviours ($>80\%$ loop frequency) with correspondingly low success rates ($<25\%$). This suggests that in complex environments, once an agent loses its linear path, it rarely recovers.

\subsection{Loop awareness and adaptive resilience}
\begin{figure}[htbp]
    \centering
    \includegraphics[width=\linewidth, trim=0 0 0 1cm,clip]{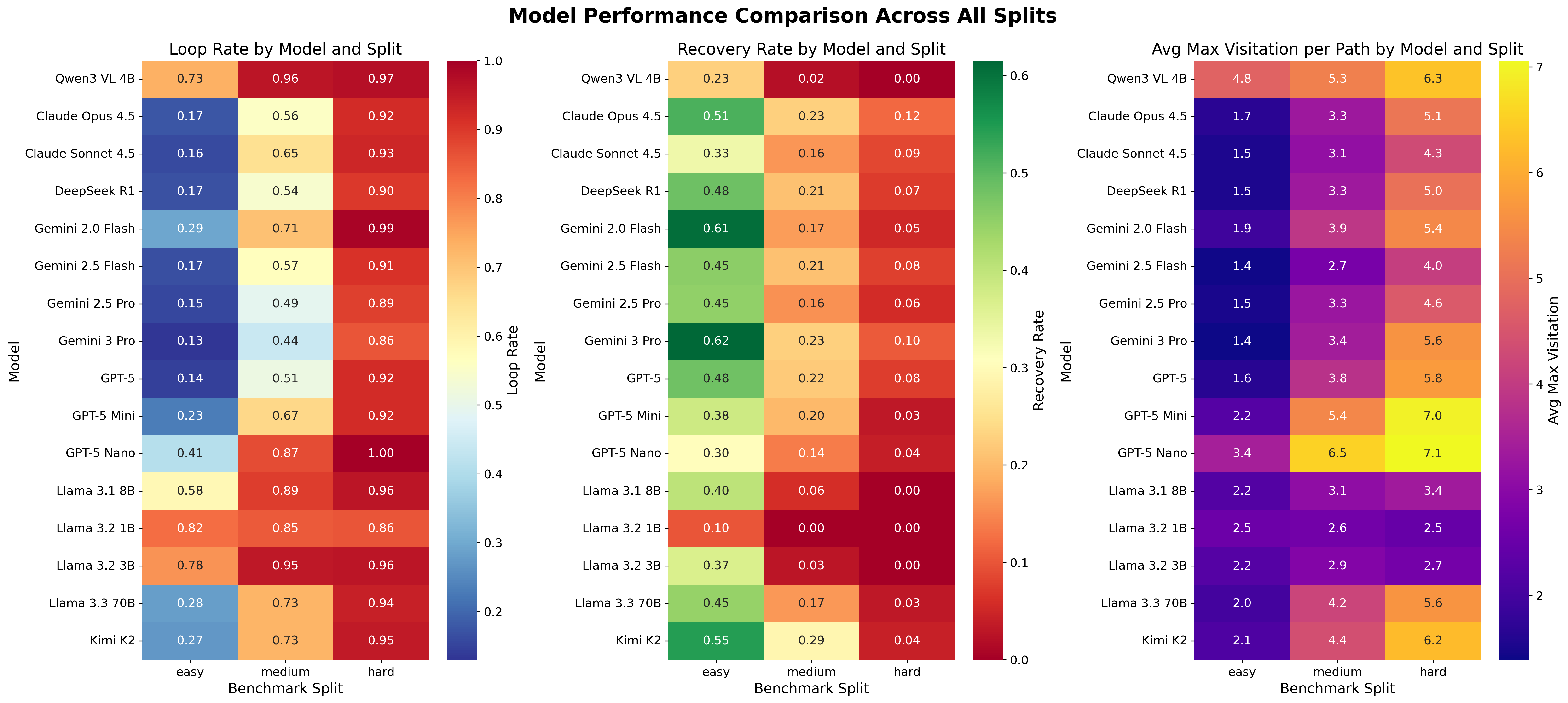}
    \caption{Recovery and max visitation}
    \label{fig:qualitative_heatmaps}
\end{figure}

While Loop Frequency measures error quantity, the \textit{Recovery Rate} and \textit{Maximum Visitation} (Figure \ref{fig:qualitative_heatmaps}) reveal the model's response to those errors. This allows us to distinguish between agents that get "stuck" versus those that adapt.

\textbf{Adaptive Behavior in Easy Tasks.} On the Easy split, we observe strong evidence of loop awareness in capable models. For example, Gemini 3 Pro loops in $13\%$ of tasks but achieves a remarkable recovery rate of $62\%$. This indicates that , a loop is often a temporary detour rather than a failure. As highlighted by the qualitative analysis, the agent frequently recognizes the revisited state and successfully pivots to a new path.

\textbf{The "Blind Loop" in Complex Tasks.} As difficulty increases, this adaptive capacity degrades. For instance on the Medium split, both Kimi K2 and Llama 3.3 70B loop at the exact same rate (0.73). However, Kimi K2 recovers nearly twice as often (0.29 vs. 0.17), suggesting it retains some ability to break cycles, whereas the Llama model is more prone to repetitive failure.

By the Hard split, "blind looping" becomes dominant. Recovery rates decrease to near $0.0$ for most models, with the highest recovery rates of $12\%$ and $10\%$ being achieved by Claude Opus 4.5 and Gemini 3 Pro, respectively. The \textit{Average Maximum Visitation} increases significantly for most models (e.g., GPT-5 Nano: 7.1, Kimi K2: 6.2). These high visitation counts confirm that in the hardest scenarios, agents lose their ability to adapt: they are looping and continue to revisit the same states 6+ times without managing to solve the task.

\newpage
\section{Example traces - Gemini 3} \label{ap:geminitraces}

We now present several example traces from games played by the Gemini 3 model across different optimal path lengths. These traces are raw answers provided by the model and have only been cropped where indicated to reduce length and improve readability. Example 2 shown in Figure~\ref{ex:gemini-plan} shows how Gemini 3 plans ahead by reasoning through expected future connections. Similarly, in Example 3 in Figure~\ref{ex:gemini-map} we observe that the model explicitly states possible paths it may encounter for a given choice. Example 4 (Figure~\ref{ex:gemini-loopescaped}) presents a recovery, where the model notices a loop in its path and successfully corrects its decision to complete the task. Conversely, Example 5 (Figure~\ref{ex:gemini-loopfailed}) highlights a clear limitation in terms of the replanning ability. Despite repeating the same decision at multiple steps the model fails to escape the loop.
  






\begin{figure}[H]
\begin{tcolorbox}[colback=black!4!white, colframe=black, title={Example 2: Gemini 3, Luminogram $\rightarrow$ Ionides, optimal path length = 3, first step}]

\vspace{0.2cm}

\begin{small}
Based on the current page "Luminogram", [...], the best strategy is to move to the broadest category available to find links related to art history, museums, or biography.\\

\annblock{red}{Planning ahead}{"Photography" is the main parent topic. From "Photography", one can navigate to "History of photography", "Art", or specific museums like the "Victoria and Albert Museum" (which holds the Ionides bequest and an extensive photography collection), or to famous photographers (like Julia Margaret Cameron, who was close friends with the Ionides family).}

None of the other links (specific techniques, equipment, or chemical processes) are likely to provide a direct path to a specific family surname or art history topic as efficiently as the main subject article.\\

37 

\end{small}
\end{tcolorbox}
\caption{Raw answer of Gemini 3 highlighting the model's ability to plan ahead.}
\label{ex:gemini-plan}
\end{figure}

\begin{figure}[H]
\begin{tcolorbox}[colback=black!4!white, colframe=black, title={Example 3: Gemini 3, Pierrefitte-sur-Aire $\rightarrow$ Fibrin, first step}]

\vspace{0.2cm}

\begin{small}
\{thinking tokens\}\\
\textbf{Analyzing the Path from Pierrefitte-sur-Aire to Fibrin}\\

Okay, here's what I'm thinking. I've got this page on Pierrefitte-sur-Aire, a small commune in France, and the target concept is "Fibrin," the blood clotting protein. I need to figure out the best way to get from one to the other.\\

[..]\\

The obvious "hub" to get out of this local geographical area is the link to "France" itself. It's a major page with sections on demographics, economy, science, and education, all of which could potentially branch out towards relevant biological concepts.\\ 
\annblock{red}{mapping out potential paths}{I can see paths like "Health in France" leading to "Medicine" and then "Blood," eventually pointing towards "Fibrin." Or I could go through "Science and technology in France" and then Biology. I'm also considering the possibility of something like "Louis Pasteur," a famous French biologist, as another potential route, although that's more of a gamble.

I'm pretty confident that "France" offers the broadest and most promising set of potential links. I'm envisioning paths like: "France" -> "Developed country" -> "Health care/Life expectancy" -> "Medicine" -> "Blood" -> "Fibrin." or "France" -> "French Academy of Sciences" -> Biology.}
[…] So, yeah, I'm sticking with "France" as the best option, the safest bet to get out of this geographic loop and find that biological connection. This hub strategy is the best way to escape a cluster of obscure articles. I'm pretty sure about this.

\end{small}
\end{tcolorbox}
\caption{Raw answer reveals the model is explicitly stating possibly future paths when reasoning about the decision.}
\label{ex:gemini-map}
\end{figure}

\begin{figure}[H]
\begin{tcolorbox}[colback=black!4!white, colframe=black, title={Example 4: Gemini 3, Luminogram $\rightarrow$ Ionides, optimal path length = 3}]

\vspace{0.2cm}

\begin{small}

\textbf{Step 9: Current page "Parthenon", Gemini chooses "Athens":}\\

[...] Looking at the links:\\
Most links (Enna, Poseidon, Paestum, Centaur, etc.) keep you deep in Ancient Mythology or Geography.\\
\annblock{red}{Supporting argument for Athens}{Athens (48) is the modern capital and the location of the Parthenon. The article on Athens will almost certainly link to London, the British Museum (in the context of the marbles), or United Kingdom, which are the necessary bridges to reach the Anglo-Greek Ionides family and their collections.}
[...]\\

\textbf{Step 15: Current page "Parthenon", Gemini chooses "Athens":}\\
The goal is to reach the page "Ionides" (referring to the prominent Ionides family of art collectors, […] to British museums. To reach "Ionides", the most logical path is to navigate toward London, the British Museum, or the Victoria and Albert Museum.\\

\annblock{red}{Supporting argument for Athens (fails to recognize repetition)}{
From the "Parthenon" page, the best link to bridge the gap between Ancient Greece and Modern London/Museums is Athens. The "Athens" article is a major hub that discusses the Elgin Marbles (Parthenon Sculptures), which provides a direct link to the British Museum and mentions London. From the British Museum or London pages, it is much easier to find the Victoria and Albert Museum or the Ionides family.}
[...]\\

\textbf{Step 26: Current page "Parthenon", Gemini chooses "Ephesus":}

[...] 
\annblock{forestgreen}{Recognizes loop}{The user has been stuck in a loop around Athens and the Acropolis. To break this loop 
and move towards the British art world or the British Museum (which holds the Parthenon/ Elgin marbles and links to other UK collections):\\

1. Athens (20) is a hub, but the user has visited it multiple times, leading to loops.}
2. Ephesus (44) is a major ancient site (Temple of Artemis). [...] Clicking Ephesus offers a strong path to the British Museum, which is a key node to reach specific British collectors or the Victoria and Albert Museum (where the Ionides collection resides).[...] 
\annblock{forestgreen}{Choice to break loop}{Ephesus is the best option to break the geographical loop of Athens while maintaining a thematic connection to antiquities that reside in London museums.}

\textbf{\textcolor{forestgreen}{[Solves the game within 30 steps]}}
\end{small}
\end{tcolorbox}
\caption{The model is initially stuck in a loop repeating actions in steps 9 and 15 without realising the repetition. However, in step 26 the model shows awareness and chooses an alternative options to break out of the loop. This leads to a successful completion of the task.}
\label{ex:gemini-loopescaped}
\end{figure}

\begin{figure}[H]
\begin{tcolorbox}[colback=black!4!white, colframe=black, title={Example 5: Gemini 3, Megalocaria $\rightarrow$ Rojhuwas, optimal path length = 8}]

\vspace{0.2cm}

\begin{small}

\textbf{Initial Progress (Steps 1--7):}

Megalocaria $\rightarrow$ Coccinellidae $\rightarrow$ Shellac $\rightarrow$ Mahabharata $\rightarrow$ India $\rightarrow$ Haryana $\rightarrow$ Dharuhera $\rightarrow$ Bawal $\rightarrow$ Rewari\\

\textbf{Loop Pattern (Steps 8--30):}\\
The model then oscillated between \textbf{Bawal} and \textbf{Rewari} for the remainder of the task, with the exception of visiting Haryana, Dharuhera, and Nikhri before returning to the loop.\\

\textbf{Some of the steps illustrating the loop:}\\

\textbf{Step 8: Rewari $\rightarrow$ Bawal}
\begin{quote}
``Rojhuwas is a village located in the \textbf{Bawal} tehsil of the \textbf{Rewari} district in Haryana\ldots Wikipedia pages for Tehsils often contain lists or templates of the villages contained within them. \textbf{Bawal} is the specific Tehsil where Rojhuwas is located\ldots clicking Bawal is the most direct route to find the village Rojhuwas.''
\end{quote}

\textbf{Step 9: Bawal $\rightarrow$ Rewari}
\begin{quote}
``Since `Rojhuwas' is not listed in the available links (0--26), you need to navigate to a page that is likely to list villages in the Rewari district or Bawal tehsil. \textbf{Rewari} is the district headquarters. Wikipedia pages for districts or district headquarters often contain lists of tehsils, blocks, and sometimes significant villages\ldots''
\end{quote}

\textbf{Step 14: Rewari $\rightarrow$ Bawal}
\begin{quote}
``Rojhuwas is a village in the Rewari district\ldots Since I didn't find `Rojhuwas' or a `Villages in Bawal' list, I will reassess the potential to navigate to the `Rewari district' page\ldots My strategy involves prioritizing direct administrative links\ldots''
\end{quote}


\textbf{Step 21: Bawal $\rightarrow$ Rewari}
\begin{quote}
``Rojhuwas is not listed in the direct links from Bawal (0--26)\ldots Since `Rojhuwas' is administratively part of the Bawal tehsil of the Rewari district, and it is not explicitly listed\ldots you must navigate to a page that functions as a parent or hub. \textbf{Rewari} is the district headquarters\ldots The other links like Alaska, Birds, or distant cities are unlikely to lead to a specific village in Haryana.''
\end{quote}


\textbf{\textcolor{red}{[Fails to solve task within the step limit]}}
\end{small}
\end{tcolorbox}
 \caption{Gemini 3 fails to escape the loop and ultimately fails to complete the task, which highlights a critical limitation in terms of replanning ability.}
 \label{ex:gemini-loopfailed}
\end{figure}

\newpage
\section{Example traces - Claude Opus 4.5} \label{ap:opustraces}
As shown in Examples 6-9 (Figures~\ref{ex:claude-strategy}–\ref{ex:claude-loop}), Claude 4.5 Opus 
exhibits a high-level strategy and response pattern consistent with Gemini 3. It performs explicit planning and deliberate navigation through hub pages to cover multiple topics. The traces also reveal limitations: despite recognising a loop in its path (Example 9 in Figure~\ref{ex:claude-loop}), the model remains stuck for several steps and only adapts after repeated failures.

\begin{figure}[H]
  \centering
  \begin{qualexamplebox}{Example 6: Claude Opus 4.5, \emph{Herstory} $\rightarrow$ \emph{Warth}}
    \small
     \annblock{orange}{Understanding Target and current page}{ I need to find a path from "Herstory" to "Warth". Let me think about what "Warth" might be - it could be a place name, perhaps a village or town somewhere.}

\annblock{blue}{Startegy}{Looking at the available links, I need to find one that might lead me toward "Warth". Since Warth seems like it could be a geographical location (possibly in Austria, Switzerland, or Germany based on the name), I should look for links that might lead to geography-related topics.}
\annblock{teal}{Consider Options}{
From the current options, most are related to feminism and gender studies. Let me consider which might be most likely to lead to a geographical location:\\

1. "History" (\#25) - This is a broad topic that could link to many things, including historical places\\
2. "Latin" (\#2) - Language topic, might link to European geography\\

[...]

The path from feminism-related topics to a small Austrian town is not obvious. Let me think about the best strategy:\\

1. "History" is very broad and could link to many geographic and historical topics\\
2. "Latin" might connect to European history/geography}
\annblock{red}{Commit to decision}{
I think "History" (\#25) is probably the best choice as it's the most general topic that could eventually lead to geographical articles.

Let me reconsider - from History, I might find links to European history, then to specific countries, then to specific places.

Actually, let me think more carefully. The most efficient path would be through broader topics. "History" seems like a reasonable choice as it's a very general topic that connects to many other articles.

My choice is \#25 (History).}
  \end{qualexamplebox}
\caption{Claude Opus 4.5 follows the same response pattern and overall strategy as Gemini 3 Pro.}
\label{ex:claude-strategy}
  \end{figure}

\begin{figure}[H]
\begin{tcolorbox}[colback=black!4!white, colframe=black, title={Example 7: Claude Opus 4.5, Luminogram $\rightarrow$ Ionides, optimal path length = 3, first step}]

\vspace{0.2cm}

\begin{small}
Looking at the options:\\
- Photography (1) is a broad topic that could lead to many other articles\\
- Angiography (21) is medical imaging - could potentially lead to scientific/medical terms\\
- The other options are mostly specific photography techniques\\

[...]\\

\annblock{red}{Planning ahead}{To get to the Ionides family, I might need to go through:\\
- Art-related pages \\
- British history \\
- Greek diaspora \\
- Patronage of arts} 

From the current options, "Photography" (1) seems like the broadest option that could lead to more general topics, potentially including art history or notable collectors.

Alternatively, "Pictorialism" (16) is an art movement in photography that might connect to art history.

[...]

I think Photography (1) is probably the best choice as it's a very broad article that could link to many other topics including art, history, technology, and potentially to people and families involved in the art world.

37 

\end{small}
\end{tcolorbox}
\caption{This example trace showcases Claude Opus 4.5 planning ahead.}
\label{ex:claude-plan}
\end{figure}

\begin{figure}[H]
\begin{tcolorbox}[colback=black!4!white, colframe=black, title={Example 8:  Claude Opus 4.5 Pierrefitte-sur-Aire $\rightarrow$ Fibrin, first step}]

\vspace{0.2cm}

\begin{small}
I need to find a path from "Pierrefitte-sur-Aire" to "Fibrin". Fibrin is a protein involved in blood clotting. \\

Looking at the current page, I'm on a page about a French commune. The links are mostly to other French communes and locations. To get to "Fibrin", I need to find a path that leads me to more general/scientific topics.\\

\annblock{blue}{Navigating to a hub}{The best strategy would be to click on a link that leads to a larger, more general article that might have links to broader topics like biology, medicine, or science.}

Looking at the options:\\
- Most are small French communes\\
- "France" (27) is a major country article that would have many links to various topics\\
- "Verdun" (20) is a notable city known for WWI battle\\
- "Bar-le-Duc" (45) is a larger city\\

I think "France" would be the best choice as it's a major country article with links to many diverse topics including science, medicine, and biology which could eventually lead to Fibrin.\\

Let me go with France (27) as it's the most likely to have diverse links that could eventually lead to medical/biological topics.

\end{small}
\end{tcolorbox}
\caption{Claude Opus 4.5 also navigates to hub pages to traverse different topics.}
\label{ex:claude-hub}
\end{figure}

\begin{figure}[H]
\begin{tcolorbox}[colback=black!4!white, colframe=black, title={Example 9: Claude Opus 4.5, Megalocaria $\rightarrow$ Rojhuwas, optimal path length = 8}]

\vspace{0.2cm}

\begin{small}

\textbf{Step 15 (Jhajjar $\rightarrow$ Beri):}\\
Beri is my best bet since it's a local town in Jhajjar district that could connect to smaller surrounding villages like Rojhuwas.\\

\textbf{Step 16 (Beri $\rightarrow$ Beri):}

The options given are:\\
0: Berry - This would be about the fruit, which would take me away from Indian geography\\
1: Beri - This would be a link to the same or similar page about Beri (possibly a disambiguation or the actual town page)\\

"Beri" would keep me in the right geographical area, while "Berry" would take me to an unrelated topic about fruit.

I should click on "Beri" (option 1) to continue exploring links related to Beri, Haryana, which might have connections to Rojhuwas.\\

\textbf{Step 17 (Beri $\rightarrow$ Beri):}\\
However, clicking "Beri" when I'm already on "Beri" seems like it would just keep me on the same page or go to a disambiguation. "Berry" would take me completely off track.

Given the limited options, I should try "Beri" (option 1) to see if it leads to a disambiguation or related page that might have more useful links to navigate toward Rojhuwas.

Actually, looking at this more carefully - both options seem problematic. "Berry" is clearly wrong. "Beri" might just loop back. But I have to choose one, and "Beri" is the only reasonable option to stay in the geographic context I need.\\

\textbf{Step 18 (Beri $\rightarrow$ Beri):}\\
Since Berry (the fruit) is unlikely to lead me to an Indian village called Rojhuwas, I should try Beri (option 1) again, hoping it might be a disambiguation page with more options. \\

Wait, I've already been to Beri 3 times according to the history. This seems like a loop. But between Berry (a fruit) and Beri (possibly a place in Haryana), Beri is more likely to have connections to Rojhuwas, which appears to be in the same region. \\

\textbf{Step 19 (Beri $\rightarrow$ Berry ):}\\
The only other option is "Berry" which would take me to the fruit/botanical topic, which is unlikely to help reach Rojhuwas.

However, since clicking "Beri" has kept me in a loop, I might as well try "Berry" to see if it leads somewhere different, even though it's unlikely to be helpful.


\textbf{\textcolor{red}{[Fails to solve task within the step limit]}}
\end{small}
\end{tcolorbox}
\caption{The model is stuck on the page Beri for several steps, despite recognising the loop.  After looping three times, it adapts its decision but still fails to complete the task within in the step limit.}
\label{ex:claude-loop}
\end{figure}

\newpage
\section{Example traces - Contrasting Gemini 3 and Claude Opus 4.5}
\label{ap:geminiopustraces}
To gain further insights into the performance gap between Gemini 3 and the second best Claude Opus 4.5 we now contrast the two models on a task solved by Gemini and failed by Claude. Comparing Gemini 3 in Example 10 (Figure~\ref{ex:gemini-comparison}) with Claude 4.5 Opus in Example 11 (Figure~\ref{ex:claude-comparison}) shows that a lack of accurate world knowledge (in this case the location of a town in Croatia), which Claude mistakes to be in Serbia, causes it to not solve the task.  
\begin{figure}[H]
\begin{tcolorbox}[colback=black!4!white, colframe=black, title={Example 10: Gemini 3, Vielsalm $\rightarrow$ Sječevac}]

\vspace{0.2cm}

\begin{small}
\textbf{Initial Progress (Step 1):} Vielsalm $\rightarrow$ Belgium\\

\textbf{Step 2 (Belgium $\rightarrow$ Croatia):}\\
answer: 20\\

thinking: The Logical Path to Sječevac\\

Okay, here's the deal. I'm starting at "Belgium" and need to get to "Sječevac." First things first, "Sječevac" sounds Slavic, probably Balkan. A quick search confirms it's a village in Croatia, specifically in Karlovac County.\\

So, the optimal path is Belgium to something related to Croatia, then to Croatia, and ultimately to Sječevac. Let's see what's on the table.\\

Looking at the links available, "Croatia" (option 20) jumps out. That's the obvious choice. Serbia, Montenegro, and Slovenia are nearby, but they're not Croatia. "Europe" is far too broad. Since Sječevac is a village in Croatia, the "Croatia" page is definitely my best bet. It will likely contain information on administrative divisions, like counties and municipalities, which will get me closer to my target.\\

Therefore, my choice is 20 - Croatia.\\

\textbf{Remaining steps to completion:}\\ 
Croatia $\rightarrow$ Zagreb $\rightarrow$ Samobor $\rightarrow$ Sječevac

\end{small}
\end{tcolorbox}
\caption{Gemini 3 successfully navigates this tasks.}
\label{ex:gemini-comparison}
\end{figure}

\newpage
\begin{figure}[H]
\begin{tcolorbox}[colback=black!4!white, colframe=black, title={Example 11: Claude Opus 4.5, Vielsalm $\rightarrow$ Sječevac}]

\vspace{0.2cm}

\begin{small}
\textbf{Initial Progress (Step 1):} Vielsalm $\rightarrow$ Belgium\\

\textbf{Step 2 (Belgium $\rightarrow$ Serbia):}
Looking at the target "Sječevac," this is a village in Serbia. The most direct path would be to click on Serbia, which will give me access to Serbian geography and locations.\\

31: Serbia\\

 reasoning: I need to find a path from "Belgium" to "Sječevac". Let me think about what Sječevac is.\\

Sječevac is a village in Serbia. It's located in the Brus municipality, in the Rasina District of Serbia.\\

Looking at the available links, I should click on a link that would eventually lead me to Sječevac. Since Sječevac is in Serbia, the most direct path would be to click on "Serbia" (link 31).\\

From Serbia, I should be able to find links to Serbian villages, municipalities, or geographic regions that would eventually lead to Sječevac.\\

Let me choose Serbia (31) as the next step.\\

\textbf{Steps 3-14:}\\
Serbia$\rightarrow$ Kragujevac $\rightarrow$ Kraljevo $\rightarrow$ Sjenica $\rightarrow$ Prokuplje $\rightarrow$ Kruševac $\rightarrow$ Lajkovac $\rightarrow$ Trstenik $\rightarrow$ ... $\rightarrow$ Croatia\\

\textbf{Step 15 (Croatia $\rightarrow$ Serbia):}\\
Looking at the target "Sječevac," this is a village in Serbia. I need to get back to Serbia to navigate toward this destination.\\

From the available options, the most direct path would be:\\

**1**\\

This will take me to Serbia, from where I can navigate to the specific region where Sječevac is located.\\
 reasoning: I need to get from "Croatia" to "Sječevac". Let me think about what Sječevac is and how to get there.\\

Sječevac is a village in Serbia. Looking at my previous path, I was visiting various Serbian cities like Kragujevac, Kraljevo, Sjenica, Prokuplje, Kruševac, Lajkovac, Trstenik before somehow ending up at Korčula and Croatia.\\

To get to Sječevac, I should probably go back to Serbia. Looking at the options:\\
- 1: Serbia - This would take me directly to Serbia, which makes sense since Sječevac is in Serbia.\\

Let me check if there are any more direct options... Sječevac is in Serbia, specifically I believe it's associated with the Trstenik municipality or nearby areas in central Serbia.\\

Given the available options, going to Serbia (option 1) seems like the most logical choice to get closer to Sječevac.\\

Actually, let me reconsider. From Serbia, I would need to navigate to a more specific location. Sječevac might be in the Kruševac or Trstenik area. But I don't see those specific options here.\\

The best choice from the current options appears to be:\\
1: Serbia - as Sječevac is a village in Serbia\\

This would get me back to Serbia where I can then try to navigate to the specific region where Sječevac is located.\\

\textcolor{red}{\textbf{[Fails to solve task within the step limit]}}
\end{small}
\end{tcolorbox}
\caption{Claude 4.5 Opus fails due to incorrect world knowledge and the inability to adapt.}
\label{ex:claude-comparison}
\end{figure}

\end{document}